\RequirePackage{fix-cm}
\documentclass[twocolumn]{svjour3}          % twocolumn
\smartqed  % flush right qed marks, e.g. at end of proof
\usepackage{graphicx}
\usepackage{cite}
\usepackage[colorlinks,citecolor={green}]{hyperref}
\usepackage{soul}
\usepackage{multirow}
\usepackage{booktabs}
\usepackage{amsmath,amssymb} % define this before the line numbering.
\usepackage{color}

\usepackage{lineno}
\usepackage{array}
\usepackage{times}
\usepackage{longtable}
\usepackage[numbers]{natbib}
\usepackage{wrapfig}
\usepackage{subcaption}
\usepackage[misc]{ifsym}

\newcommand{\eg}{\emph{e.g. }}
\newcommand{\ie}{\emph{i.e. }}
\newcommand{\newcontent}[1]{\textcolor{black}{#1}}

\newcommand{\revision}[1]{\textcolor{black}{#1}}

\begin{document}

\title{Decoupled Classification Refinement:\\Hard False Positive Suppression for Object Detection%\thanks{Grants or other notes
}
% \subtitle{Do you have a subtitle?\\ If so, write it here}


\author{Bowen~Cheng$^{1}$         \and
        Yunchao~Wei$^{2}$         \and
        Rogerio~Feris$^{3}$         \and
        Jinjun~Xiong$^{3}$        \and
        Wen-mei Hwu$^{1}$  \and
        Thomas~S.~Huang$^{1}$ \and %etc.
        Humphrey~Shi$^{1,4}$
}

\authorrunning{Bowen Cheng \emph{et al.}} % if too long for running head

\institute{ Bowen Cheng (bcheng9@illinois.edu) \\
        Yunchao Wei (wychao1987@gmail.com) \\
        Rogerio Feris (rsferis@us.ibm.com) \\
        Jinjun Xiong (jinjun@us.ibm.com) \\
        Wen-mei Hwu (w-hwu@illinois.edu)
        \\
        Thomas S. Huang (t-huang1@illinois.edu) \\
        $\textrm{\Letter}$ Humphrey Shi (hshi10@illinois.edu) \\
       1 University of Illinois at Urbana-Champaign, USA \\
       2 University of Technology Sydney, Australia \\
       3 IBM Research, USA \\
       4 University of Oregon, USA
 }

\date{Received: date / Accepted: date}

\maketitle

\begin{abstract}
In this paper, we analyze failure cases of state-of-the-art detectors and observe that most \textit{hard false positives} result from classification instead of localization and they have a large negative impact on the performance of object detectors. We conjecture there are three factors \revision{and prove our hypothesis with experiments}:
(1) Shared feature representation is not optimal due to the mismatched goals of feature learning for classification and localization;
(2) large receptive field for different scales leads to redundant context information for small objects;
(3) multi-task learning helps, yet optimization of the multi-task loss may result in sub-optimal for individual tasks.
We demonstrate the potential of detector classification power by a simple, effective, and widely-applicable \textit{Decoupled Classification Refinement} (DCR) network. \newcontent{In particular, DCR places a separate classification network in parallel with the localization network (base detector). With ROI Pooling placed on the early stage of the classification network, we enforce an adaptive receptive field in DCR. During training, DCR samples hard false positives from the base detector and trains a strong classifier to refine classification results. During testing, DCR refines all boxes from the base detector. Experiments show competitive results on PASCAL VOC and COCO without any bells and whistles. Our codes are available at:
\href{https://github.com/SHI-Labs/Decoupled-Classification-Refinement}{this link}
}.

\keywords{Object detection \and False positive \and Error analysis \and Deep learning}

\end{abstract}

\section{Introduction}
\begin{figure*}[t]
	\centering
	\includegraphics[width=1.0\textwidth]{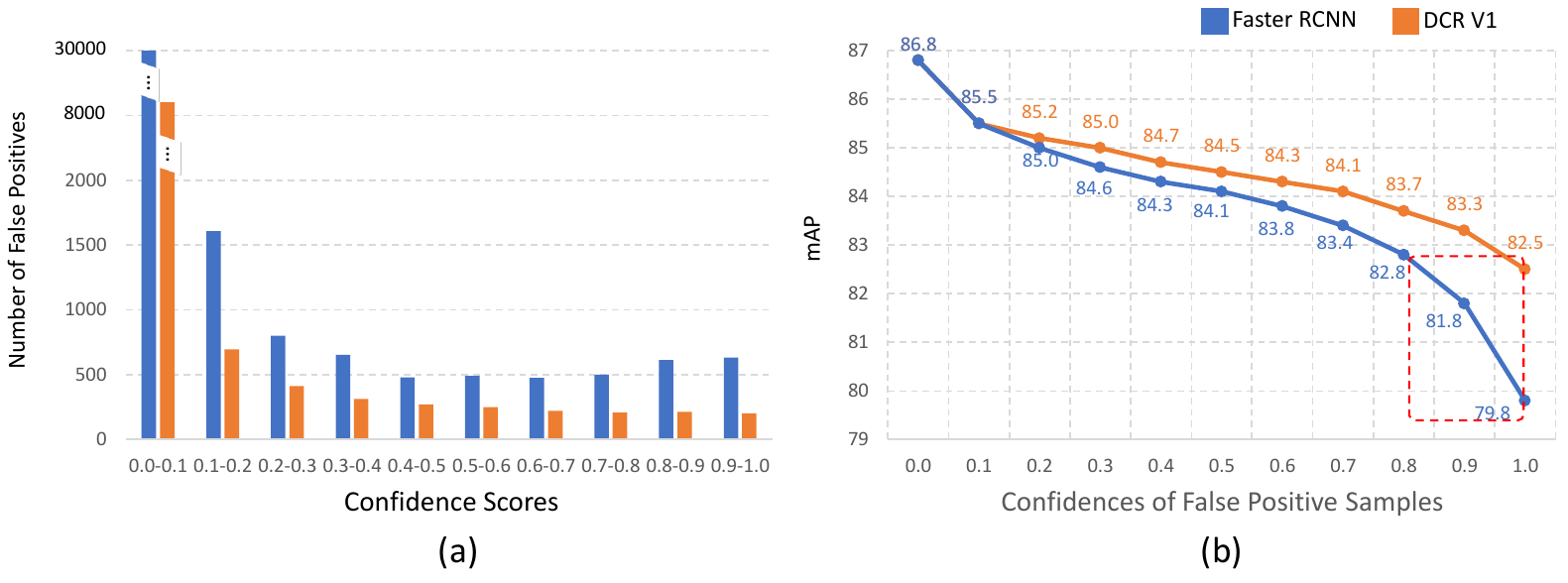}
	\caption{(a) Comparison of the number of false positives in different ranges. (b) Comparison of the mAP gains by progressively removing false positives; from right to left, the detector is performing better as false positives are removed according to their confidence scores.}
	\label{fig:motivation}
    %  \vspace{-4mm}
\end{figure*}

\begin{figure*}[t]
	\centering
	\includegraphics[width=1\textwidth]{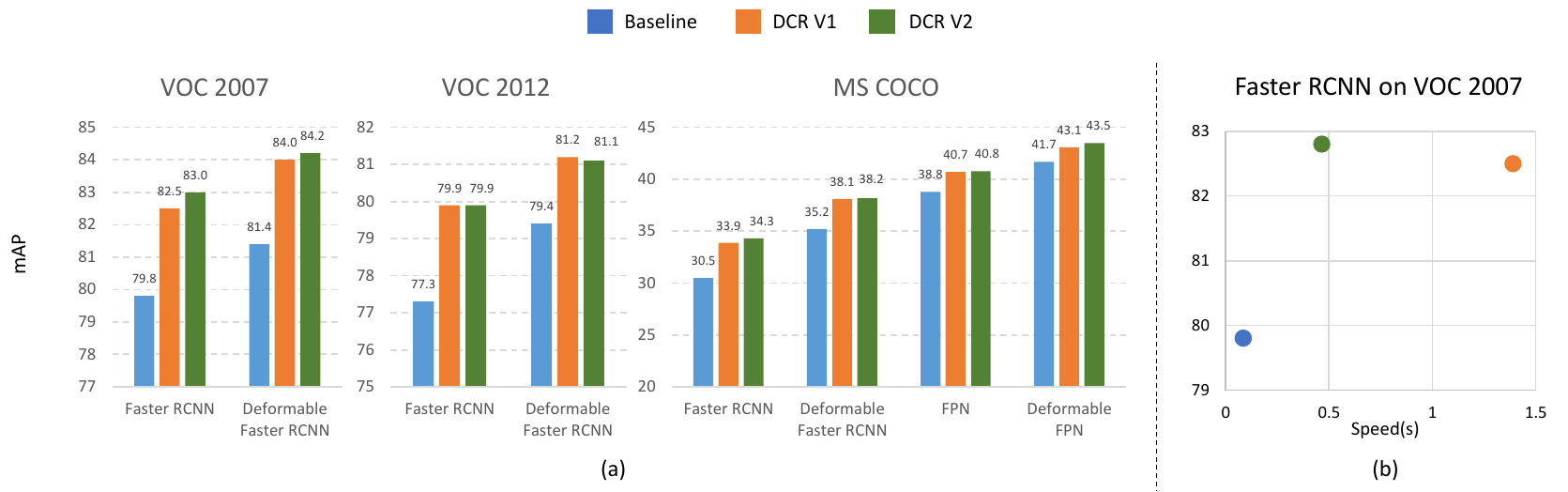}
	\caption{\newcontent{(a) Comparison of our proposed DCR V1 and V2 with baseline in terms of different Faster RCNN series and benchmarks. (b) Speed-Accuracy comparison of DCR V1 and V2 upon Faster RCNN.}}
	\label{fig:improvement}
    %  \vspace{-4mm}
\end{figure*}

Region-based approaches with convolutional neural networks (CNNs) \citep{girshick2014rich,girshick2015fast,ren2015faster,cai2017cascade,xu2017deep,li2017attentive,li2018multistage,li2017perceptual,yu2016unitbox,hu2017relation,shen2017improving,wei2018ts2c} have achieved great success in object detection. Such detectors are usually built with separate classification and localization branches on top of shared feature extraction networks, and trained with multi-task loss. In particular, Faster RCNN \cite{ren2015faster} learns one of the first end-to-end two-stage detector with remarkable efficiency and accuracy. Many follow-up works, such as R-FCN \cite{dai2016r}, Feature Pyramid Networks (FPN) \cite{lin2017feature}, Deformable ConvNets (DCN) \cite{dai2017deformable}, have been leading popular detection benchmark in PASCAL VOC \cite{Everingham10} and COCO \cite{lin2014microsoft} datasets in terms of accuracy. Yet, few work has been proposed to study what is the full potential of the classification power in Faster RCNN styled detectors.

To answer this question, in this paper, we begin with investigating the key factors affecting the performance of Faster RCNN. As shown in Fig~\ref{fig:motivation} (a), we conduct object detection on PASCAL VOC 2007 using Faster RCNN and count the number of false positive detections in different confidence score intervals (blue). Although only a small percentage of all false positives are predicted with high confidence scores, these samples lead to a significant performance drop in mean average precision (mAP). In particular, we perform an analysis of potential gains in mAP using Faster RCNN: As illustrated in Fig~\ref{fig:motivation} (b), given the detection results from Faster RCNN and a confidence score threshold, we assume that all false positives with predicted confidence score above that threshold were classified correctly and we report the correspondent hypothesized mAP. It is evident that by correcting all false positives, Faster RCNN could, hypothetically, have achieved $86.8\%$ in mAP instead of $79.8\%$. Moreover, even if we only eliminate false positives with high confidences, as indicated in the red box, we can still improve the detection performance significantly by $3.0\%$ mAP, which is a desired yet hard-to-obtain boost for modern object detection systems. 

The above observation motivates our work to alleviate the burden of false positives and improve the classification power of Faster RCNN based detectors.
By scrutinizing the false positives produced by Faster RCNN, we conjecture that such errors are mainly due to \revision{three} reasons: (1) Shared feature representation for both classification and localization may not be optimal for region proposal classification, the mismatched goals in feature learning lead to the reduced classification power of Faster RCNN; (2) Receptive fields in deep CNNs such as ResNet-101 \cite{he2016deep} are large, the whole image are usually fully covered for any given region proposals. Such large receptive fields could lead to inferior classification capacity by introducing redundant context information for small objects; and (3) Multi-task learning in general helps to improve the performance of object detectors as shown in Fast RCNN~\cite{girshick2015fast} and Faster RCNN~\cite{ren2015faster}, but the joint optimization also leads to possible sub-optimal to balance the goals of multiple tasks and could not directly utilize the full potential on individual tasks.

Following the above arguments, we propose a simple yet effective approach, named Decoupled Classification Refinement (DCR), to eliminate high-scored false positives and improve the region proposal classification results. DCR decouples the classification and localization tasks in Faster RCNN styled detectors. It takes input from a base detector, \eg the Faster RCNN, and refine the classification results using a separate classification network which does not share features with the base detector. DCR samples \textit{hard false positives}, namely the false positives with high confidence scores, from the base classifier, and then trains a stronger correctional classifier for the classification refinement. 
\revision{DCR is originally designed to not share any parameters with the Faster RCNN, however, we observe that sharing low-level features could speed up inference time without lossing accuracy.}
% Designedly, we do not share any parameters between the Faster RCNN and our DCR module, so that the DCR module can not only utilize the multi-task learning improved results from region proposal networks (RPN) and bounding box regression tasks, but also better optimize the newly introduced module to address the challenging classification cases.

\newcontent{Experimental results show the benefit of decoupling classification and localization tasks in object detection and, more interestingly, we find a new speed-accuracy trade-off in object detection which is controlled by the amount of features shared between classification network and localization network. Namely, we find that the less features shared between classification and localization task, the better performance a detector will achieve and, in the mean while, the slower the detector is during inference time. We hope this new trade-off can motivate the society to find new directions of improving over current objection architectures.}

We conduct extensive experiments based on different Faster RCNN styled detectors (\ie Faster RCNN, Deformable ConvNets, FPN) and benchmarks (\ie PASCAL VOC 2007 \& 2012, COCO) to demonstrate the effectiveness of our proposed simple solution in enhancing the detection performance by alleviating hard false positives. 
As shown in Fig~\ref{fig:motivation} (a), our approach can significantly reduce the number of hard false positives and boost the detection performance by $2.7\%$ in mAP on PASCAL VOC 2007 over a strong baseline as indicated in Fig~\ref{fig:motivation} (b).
All of our experiment results demonstrate that our proposed DCR module can provide consistent improvements over various detection baselines, as shown in Fig~\ref{fig:improvement} (a). Our contributions are fourfold:
% \vspace{-2mm}
\begin{enumerate}
\item
We analyze the error modes of region-based object detectors and formulate the hypotheses that might cause these failure cases.
\item
We propose a set of design principles to improve the classification power of Faster RCNN styled object detectors along with the DCR module based on the proposed design principles.
\item
\newcontent{We are the first to show decoupling classification and localization helps object detection task and observe a new speed and accuracy trade-off in object detection.}
\item
Our DCR modules consistently bring significant performance boost to strong object detection systems on popular benchmarks and achieve competitive results. In particular, following common practice of using ResNet-101 as backbone without any bells and whistles, we achieve mAP of $84.2\%$ and $81.2\%$ on the classic PASCAL VOC 2007 and 2012 set, respectively, and $43.5\%$ on the more challenging COCO \emph{test-dev} set.
%We show how to apply the DCR module to refine object detector and observed consistent improvement over several challenge benchmarks. We achieve (percent) the new state-of-the-art result on Pascal VOC 2007 dataset using only VOC training images and achieve a consistently (percent) improvement on COCO dataset.
\end{enumerate}
% \vspace{-6mm}

\newcontent{This paper is an extended version of our previous work \cite{cheng2018revisiting}. In particular, we make three methodological improvements: the first one is that we propose a new DCR module that can be end-to-end trained together with the base object detector; the second one is that the proposed DCR module reduces the computation significantly without sacrificing accuracy by sharing some of the computation of regional feature at the image level; and the third one is that we come up with a inference technique that can further reduces the inference time while keeping a good speed-accuracy trade-off (Fig~\ref{fig:improvement} (b)). In addition, we provide more results to demonstrate the effectiveness of our proposed method and we also perform more experiments to support our hypothesis. To justify decoupling features is important, we perform experiments to share different amount of features between classification and localization network (Table \ref{dcrv2_ablation} (a)) and we observe that the less features shared the better the performance is (mAP increases from 78.4\% to 83.0\%). We further justify the importance of adaptive receptive field by placing ROI Pooling to different stages (Table \ref{dcrv2_ablation} (b)) and we observe that the earlier ROI Pooling is placed the more adjustable receptive field can be and the better performance we get (mAP increases from 79.8\% to 83.0\%). These experiments also give new speed-accuracy trade-offs for object detectors.}

\newcontent{The rest of the paper is organized as follows. In Section~\ref{related}, some related works are introduced. In Section~\ref{problems}, we analyze problems of current object detectors. In Section~\ref{DCR}, the details of the proposed method are described. In Section~\ref{exps}, we extensively perform experiments and analysis. Finally, we conclude this paper in Section~\ref{conclusion}.}

\section{Related Works}
\begin{figure*}[ht]
	\centering
	\includegraphics[width=0.8\textwidth]{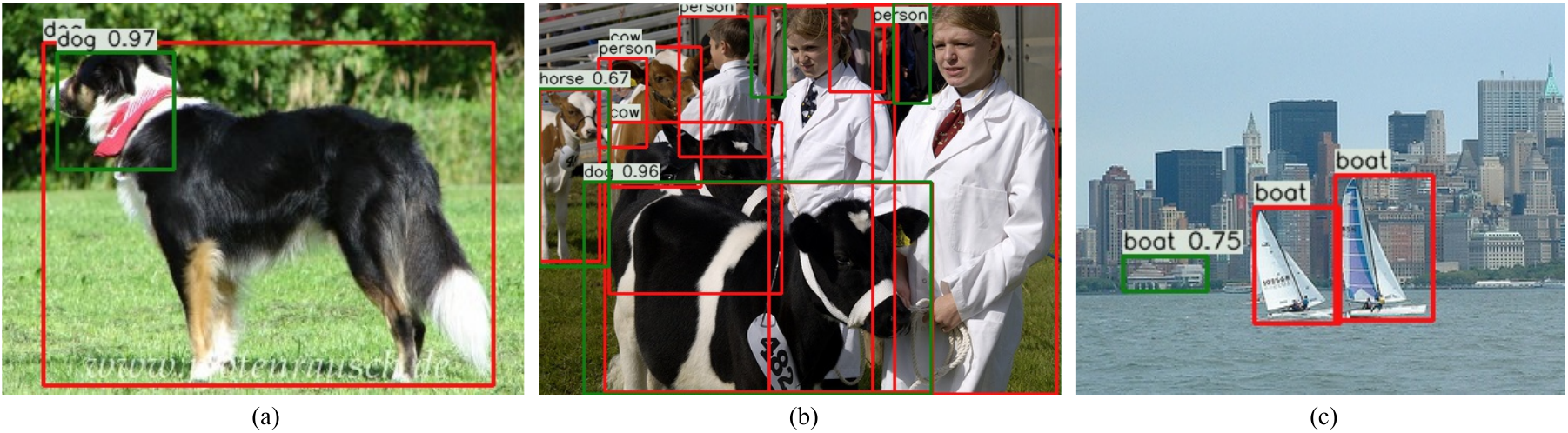}
	\caption{Demonstration of hard false positives. Results are generate by Faster RCNN with 2 fully connected layer (2fc) as detector head \cite{ren2015faster,lin2017feature}, red boxes are ground truth, green boxes are hard false positives with scores higher than 0.3; (a) boxes covering only part of objects with high confidences; (b) incorrect classification due to similar objects; (c) misclassified backgrounds.}
	\label{fig:FP}
    %  \vspace{-4mm}
\end{figure*}
\label{related}
% \vspace{-2mm}
% \subsection{Object Detection}
\noindent\textbf{Object Detection.} 
Recent CNN based object detectors can generally be categorized into two-stage and single stage. One of the first two-stage detector is RCNN \cite{girshick2014rich}, where selective search \cite{uijlings2013selective} is used to generate a set of region proposals for object candidates, then a deep neural network to extract feature vector of each region followed by SVM classifiers. SPPNet \cite{he2014spatial} improves the efficiency of RCNN by sharing feature extraction stage and use spatial pyramid pooling to extract fixed length feature for each proposal. Fast RCNN \cite{girshick2015fast} improves over SPPNet by introducing an differentiable ROI Pooling operation to train the network end-to-end. Faster RCNN \cite{ren2015faster} embeds the region proposal step into a Region Proposal Network (RPN) that further reduce the proposal generation time. R-FCN \cite{dai2016r} proposed a position sensitive ROI Pooling (PSROI Pooling) that can share computation among classification branch and bounding box regression branch. Deformable ConvNets (DCN) \cite{dai2017deformable} further add deformable convolutions and deformable ROI Pooling operations, that use learned offsets to adjust position of each sampling bin in naive convolutions and ROI Pooling, to Faster RCNN. Feature Pyramid Networks (FPN) \cite{lin2017feature} add a top-down path with lateral connections to build a pyramid of features with different resolutions and attach detection heads to each level of the feature pyramid for making prediction. Finer feature maps are more useful for detecting small objects and thus a significant boost in small object detection is observed with FPN. Most of the current state-of-the-art object detectors are two-stage detectors based of Faster RCNN, because two-stage object detectors produce more accurate results and are easier to optimize. However, two-stage detectors are slow in speed and require very large input sizes due to the ROI Pooling operation. Aimed at achieving real time object detectors, one-stage method, such as OverFeat \cite{sermanet2013overfeat}, SSD \cite{liu2016ssd,fu2017dssd} and YOLO \cite{redmon2016you,redmon2017yolo9000}, predict object classes and locations directly. Though single stage methods are much faster than two-stage methods, their results are inferior and they need more extra data and extensive data augmentation to get better results. Recently, there advances a new group of one-stage object detectors named anchor-free methods. One type of anchor-free methods~\cite{Law_2018_ECCV,zhou2019bottom,duan2019centernet,zhou2019objects,cheng2019panoptic_1,cheng2019panoptic_2} are motivated by human pose estimation~\cite{cao2017realtime,papandreou2017towards,cheng2019bottom}. They first define ``keypoints'' for an object (\eg center point~\cite{zhou2019objects,cheng2019panoptic_1,cheng2019panoptic_2}, corner point~\cite{Law_2018_ECCV} or extreme point~\cite{zhou2019bottom}) and regress to the object bounding box. Another type of anchor-free methods~\cite{tian2019fcos,zhu2019feature} are motivated by fully convolutional networks (FCN) for semantic segmentation~\cite{long2015fully,deeplabv3plus2018,cheng2019spgnet}. They directly predict the bounding box on every pixel that belongs to an instance. Our paper follows the method of two-stage detectors \cite{girshick2014rich,girshick2015fast,ren2015faster}, but with a main focus on analyzing reasons why detectors make mistakes.
% \vspace{-6mm}

% \subsection{Classifier Cascade}
\noindent\textbf{Classifier Cascade.} 
The method of classifier cascade commonly trains a stage classifier using misclassified examples from a previous classifier. This has been used a lot for object detection in the past. The Viola Jones Algorithm \cite{viola2004robust} for face detection used a hard cascades by Adaboost \cite{freund1997decision}, where a strong region classifier is built with cascade of many weak classifier focusing attentions on different features and if any of the weak classifier rejects the window, there will be no more process. Soft cascades \cite{bourdev2005robust} improved \cite{viola2004robust} built each weak classifier based on the output of all previous classifiers. Deformable Part Model (DPM) \cite{felzenszwalb2010object} used a cascade of parts method where a root filter on coarse feature covering the entire object is combined with some part filters on fine feature with greater localization accuracy. More recently, Li et al. \cite{li2015convolutional} proposed the Convolutional Neural Network Cascade for fast face detection. Our paper proposed a method similar to the classifier cascade idea, however, they are different in the following aspects. The classifier cascade aims at producing an efficient classifier (mainly in speed) by cascade weak but fast classifiers and the weak classifiers are used to reject examples. In comparison, our method aims at improving the overall system accuracy, where exactly two strong classifiers are cascaded and they work together to make more accurate predictions. More recent Cascade RCNN \cite{cai2017cascade} proposes training object detector in a cascade manner with gradually increased IoU threshold to assign ground truth labels to align the testing metric \ie average mAP with IOU 0.5:0.05:0.95.

% \vspace{-4mm}

%  DCN producing better results by stronger localization powers.
%, where a CNN classifier with very small input size of $12\times12$ is first used to reject 90\% of the detection windows followed with bounding box regression with $12\times12$ boxes as input, a cascaded CNN classifier with input size $24\times24$ is used to further reject 90\% of the remaining windows followed with another bounding box regression network with $24\times24$ input

\section{Problems with Faster RCNN}
\label{problems}

\revision{We observe Faster RCNN produces 3 typical types of hard false positives (which may be the case for any object detector), }
%Faster RCNN produces 3 typical types of hard false positives, 
as shown in Fig \ref{fig:FP}:
(1) The classification is correct but the overlap between the predicted box and ground truth has low IoU, \eg $<0.5$ in Fig \ref{fig:FP} (a). This type of false negative boxes usually cover the most discriminative part  and have enough information to predict the correct classes due to translation invariance. 
(2) Incorrect classification for predicted boxes but the IoU with ground truth are large enough, \eg in Fig \ref{fig:FP} (b). It happens mainly because some classes share similar discriminative parts and the predicted box does not align well with the true object and happens to cover only the discriminative parts of confusion. Another reason is that the classifier used in the detector is not strong enough to distinguish between two similar classes. 
(3) the detection is a ``confident" background, meaning that there is no intersection or small intersection with ground truth box but classifier's confidence score is large, \eg in Fig \ref{fig:FP} (c). Most of the background pattern in this case is similar to its predicted class and the classifier is too weak to distinguish. Another reason for this case is that the receptive field is fixed and it is too large for some box that it covers the actual object in its receptive field. In Fig \ref{fig:FP} (c), the misclassified background is close to a ground truth box (the left boat), and the large receptive field (covers more than 1000 pixels in ResNet-101) might ``sees'' too much object features to make the wrong prediction. Given above analysis, we can conclude that the hard false positives are mainly caused by the suboptimal classifier embedded in the detector. The reasons may be that: (1) feature sharing between classification and localization, (2) detector's receptive field does not change according to the size of objects and (3) optimizing the sum of classification loss and localization loss.

% \vspace{-2mm}
\subsection{Problem with Feature Sharing}
\label{problem1}
Detector backbones are usually adapted from image classification model and pre-trained on large image classification dataset. These backbones are original designed to learn scale invariant features for classification. Scale invariance is achieved by adding sub-sampling layers, \eg max pooling, and data augmentation, \eg random crop. Detectors place a classification branch and localization branch on top of the same backbone, however, classification needs \textbf{translation invariant} feature whereas localization needs \textbf{translation covariant} feature. During fine-tuning, the localization branch will force the backbone to gradually learn translation covariant feature, which might potentially down-grade the performance of classifier.

\revision{Ablation studies (Table~\ref{dcrv2_ablation}) show that sharing less features between classification and localization branch indeed is helpful which support our hypothesis on feature sharing. }

% \vspace{-6mm}
\subsection{Problem with Receptive Field}
\label{problem3}
Deep convolutional neural networks have fixed receptive fields. For image classification, inputs are usually cropped and resized to have fixed sizes, \eg $224 \times 224$, and network is designed to have a receptive field little larger than the input region. However, since contexts are cropped and objects with different scales are resized, the ``effective receptive field'' is covering the whole object.

Unlike image classification task where a single large object is in the center of a image, objects in detection task have various sizes over arbitrary locations. In Faster RCNN, the ROI pooling is introduced to crop object from 2-D convolutional feature maps to a 1-D fixed size representation for the following classification, which results in fixed receptive field (\ie the network is attending to a fixed-size window of the input image). In such a case, objects have various sizes and the fixed receptive field will introduce different amount of context. For a small object, the context might be too large to focus on the object whereas for a large object, the receptive field might be too small that the network is looking at part of the object. Although some works introduce multi-scale features by aggregating features with different receptive field, the number of sizes is still too small comparing with the number various sizes of objects.

\revision{Recently, Deformable ConvNets V2 (DCN V2)~\cite{zhu2019deformable} provides a deeper analysis on the spatial support of regular convolution and deformable convolution and the authors find some interesting results that can support our hypothesis. The authors visualize the error-bounded saliency regions which can be interpreted as approximately the size of receptive field. DCN V2 observes that the receptive field of deformable convolution is changing according to the scale of objects. That is, feature within a smaller object has smaller receptive field and feature within larger object has larger receptive field. Furthermore, the receptive field of features within an object only attend to the object region. In~\cite{zhu2019deformable}, deformable convolution learns to adapt its receptive field to mainly cover the object without introducing any background context. In contrast, the receptive field of regular convolution does not vary much and it covers much background context for small object while it only covers partial region for large objects. This observation together with the fact that deformable convolution performs better than regular convolution in Faster RCNN support our hypothesis that ``adaptive'' receptive field is necessary.  }

% \vspace{-3mm}
% \vspace{-4mm}
\subsection{Problem with Optimization}
\label{problem2}

Faster RCNN series are built with a feature extractor as backbone and two task-specified branches for classifying regions and the other for localizing correct locations. Denote loss functions for classification and localization as $L_{cls}$ and $L_{bbox}$, respectively. Then, the optimization of Faster RCNN series is to address a Multi-Task Learning (MTL) problem by minimizing the sum of two loss functions: $L_{detection}=L_{cls}+L_{bbox}$. \revision{However, MTL for object detection is not studied under the recent powerful classification backbones and it may be suboptimal for stronger backbones. Although it is hard to provide theoretical proof for the assumption of sub-optimal optimization in multi-task learning, some experiments results show that not using MTL may achieve better results. The purpose of this assumption is not to make any conclusion, but to motivate the society to further study multi-task learning in object detection.}

% \section{Revisiting RCNN for Improving Faster RCNN}
% \vspace{-4mm}
\section{\newcontent{Decoupled Classification Refinement}}
% \vspace{-2mm}
\label{DCR}

\begin{figure*}[t]
% 	\vspace{-6mm}
	\centering
	\includegraphics[width=0.85\textwidth]{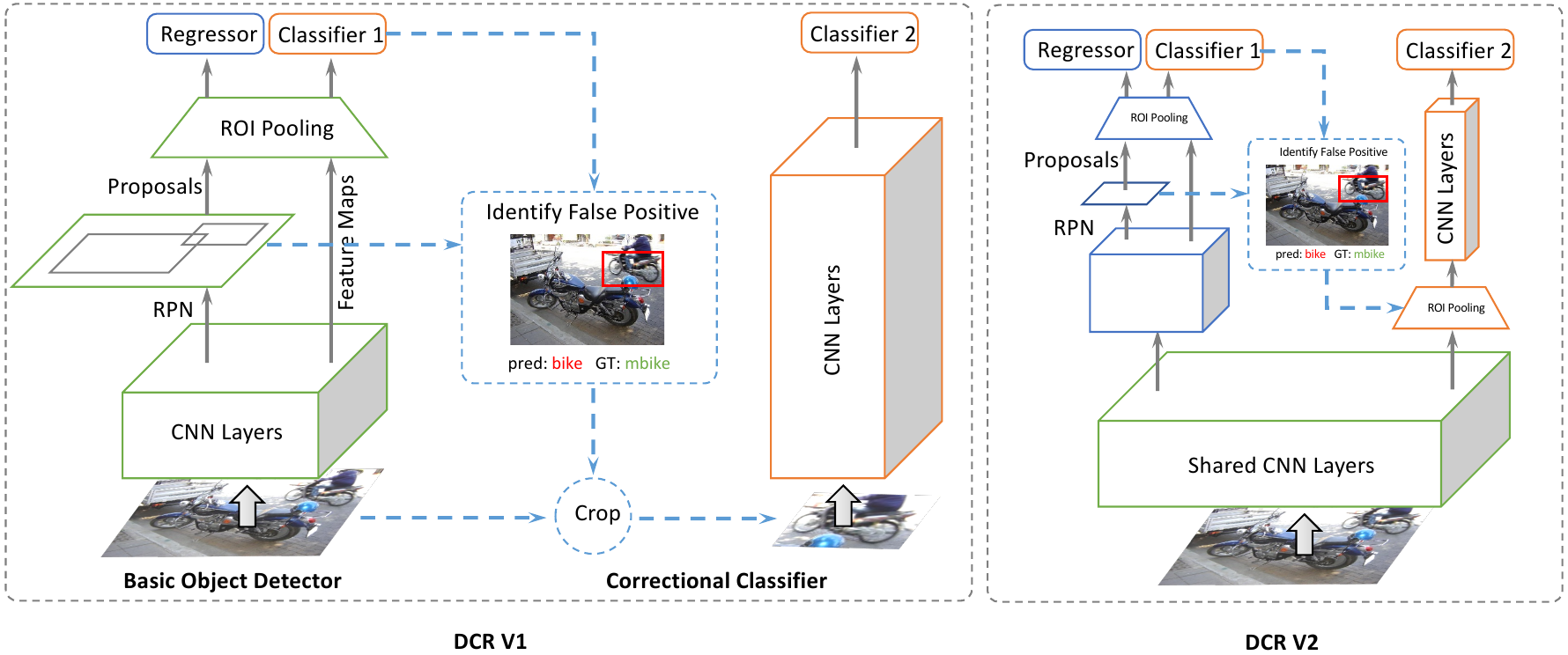}
	\caption{\newcontent{Left: DCR V1 module \cite{cheng2018revisiting}. Right: our proposed DCR V2 module.}}
	\label{fig:framework}
% 	\vspace{-4mm}
\end{figure*}

In this section, we look back closely into the classic RCNN \cite{girshick2014rich} method, and give an in-depth analysis of why RCNN can be used as a ``complement'' to improve Faster RCNN. Based on our findings, we provide a simple yet effective decoupled classification refinement (DCR) module, that can be easily added to any current state-of-the-art object detectors to provide performance improvements.
% \vspace{-4mm}
\subsection{Learning from RCNN Design}
% \vspace{-1mm}

We train a modified RCNN with ResNet-50 as backbone and Faster RCNN predictions as region proposals. We find that with RCNN along, the detection result is deteriorated. Since RCNN does not modify box coordinate, the inferior result means worse classification. We find that many boxes having small intersections with an object are classified as that object instead of the background which Faster RCNN predicts. Based on this finding, we hypothesize that the drawback of RCNN is mainly root from that classification model is pre-trained without awaring object location. Since ResNet-50 is trained to be translation-invariance on ImageNet in multi-crop manner, no matter how much the intersection of the crop to the object is, classifier is encouraged to predict that class. This leads to the classifier in RCNN being ``too strong'' for proposal classification, and this is why RCNN needs a carefully tuned sampling strategy, \ie a ratio of 1:3 of \emph{fg} to \emph{bg}. Straightforwardly, we are interested whether RCNN is ``strong'' enough to correct hard negatives. We make a minor modification to multiply RCNN classification score with Faster RCNN classification score and observe a boost of 1.9\% (from 79.8\% to 81.7\%)! Thus, we consider that RCNN can be seen as a compliment of Faster RCNN in the following sense: the classifier of Faster RCNN is weaker but aware of object location whereas the classifier of RCNN is unaware of object location but stronger. Based on our findings, we propose the following three principals to design a better object detector.

% \subsection{Design Principals}
% \vspace{-3mm}
\subsubsection{Decoupled Features}

Current detectors still place classification network and localization network on the same backbone, hence we propose that classification head and localization head should not share parameter (as the analysis given in Section \ref{problem1}), resulted in a decoupled feature using pattern by RCNN.

\newcontent{To demonstrate it is necessary to decouple classification and localization network, we explore sharing different amount of features between classification and localization. More specifically, we use two same networks, one for classification and one for localization and experiment with the number of stages that are shared (a ``stage'' is a group of features that have the same resolution). Results are shown in Table~\ref{dcrv2_ablation} (a). If all features are shared, the performance is 78.4\% mAP. When we share less features, the performance monotonically increases to 83.0\% when only the first stage is shared.}

% \vspace{-3mm}
% \subsubsection{Decoupled Optimization}
% RCNN also decouples the optimization for object proposal and classification. 
% In this paper, we make a small change in optimization. We propose a novel two-stage training where, instead of optimizing the sum of classification and localization loss, we optimize the concatenation of classification and localization loss, $L_{detection} = [L_{cls} + L_{bbox}, L_{cls}]$, where each entry is being optimized independently in two steps.

\subsubsection{Adaptive Receptive Field}
The most important advantage of RCNN is that its receptive field always covers the whole ROI, 
\ie the receptive field size adjusts according to the size of the object by cropping and resizing each proposal to a fixed size. We agree that context information may be important for precise detection, however, we conjuncture that different amount of context introduced by fixed receptive field might cause different performance to different sizes of objects. It leads to our last proposed principal that a detector should an adaptive receptive field that can change according to the size of objects it attends to. In this principal, the context introduced for each object should be proportional to its size, but how to decide the amount of context still remains an open question to be studied in the future. Another advantage of adaptive receptive field is that its features are well aligned to objects. Current detectors make predictions at hight-level, coarse feature maps usually having a large stride, \eg a stride of 16 or 32 is used in Faster RCNN, due to sub-sampling operations. The sub-sampling introduces unaligned features, \eg one cell shift on a feature map of stride 32 leads to 32 pixels shift on the image, and defects the predictions. With adaptive receptive field, the detector always attends to the entire object resulting in an aligned feature to make predictions. RCNN gives us a simple way to achieve adaptive receptive field, but how to find a more efficient way to do so remains an interesting problem needs studying.

\newcontent{To verify the importance of adaptive receptive field, we perform experiments by adding ROI Pooling to different stages in the parallel classification network. Results are shown in Table~\ref{dcrv2_ablation} (b). When we place the ROI Pooling at last stage, the network is less likely to adjust receptive field according to the ROI size and the performance is only 79.8\%. However, as we place the ROI Pooling to earlier stages, the performance increases monotonically to 83.0\%.}

% \vspace{-4mm}
\subsection{\newcontent{DCR V1: A Na\"ive Method}}

Following these principals, we propose a na\"ive module \newcontent{(DCR V1)} that can be easily augmented to Faster RCNN as well as any object detector to build a stronger detector. The overall pipeline is shown in Fig \ref{fig:framework} (left). The green part and the orange part are the original Faster RCNN and our proposed DCR module, respectively. In particular, DCR \newcontent{V1} mainly consists a crop-resize layer and a strong classifier. The crop-resize layer takes two inputs, the original image and boxes produced by Faster RCNN, crops boxes on the original image and feeds them to the strong classifier after resizing them to a predefined size. Region scores of DCR \newcontent{V1} (Classifier 2) is aggregated with region scores of Faster RCNN (Classifier 1) by element-wise product to form the final score of each region. \newcontent{\textbf{In DCR V1, the two parts are trained separately and the scores are only combined during test time.}}

The classification network in DCR \newcontent{V1} does not share any features with the detector backbone in order to preserve the quality of classification-aimed translation invariance feature. Furthermore, there is no error propagation between the classification network in DCR \newcontent{V1} and the base detector, thus the optimization of one loss does not affect the other. This in turn results in a decoupled pattern where the base detector is focused more on localization (than DCR V1) whereas the DCR \newcontent{V1} focuses more on classification. DCR \newcontent{V1} introduces adaptive receptive field by resizing boxes to a predefined size. Noticed that this processing is very similar to moving an ROI Pooling from final feature maps to the image, however, it is quite different than doing ROI Pooling on feature maps. Even though the final output feature map sizes are the same, features from ROI Pooling sees larger region because objects embedded in an image has richer context. We truncated the context by cropping objects directly on the image and the network cannot see context outside object regions.

% \vspace{-4mm}
\subsection{\newcontent{DCR V2: A Faster DCR Module}}

\newcontent{Although DCR V1 solves the problem of hard false positives, it also introduces extra computation overhead, including:
% \vspace{-2mm}
\begin{enumerate}
	\item Cropping and resizing a large number of boxes on the original image.
	\item Forward a large batch (usually 300 in Faster RCNN) of images to a deep network (a 152-layer ResNet).
\end{enumerate}}
% \vspace{-2mm}

\newcontent{The above computation overhead causes the real run time of DCR V1 module (1.3900 seconds/image) to be more than 100 times of that of the original Faster RCNN (0.0855 seconds/images). This number does not count the cropping and resizing time, which takes around 1$\sim$2 seconds per image with a  sequential CPU implementation.}

% \vspace{-1mm}
\newcontent{Inspired by \cite{he2014spatial,girshick2015fast}, we design a faster DCR module (DCR V2) that alleviates the computation overhead of DCR V1 module, shown in Fig \ref{fig:framework} (right). That is, we solve (1) by using a highly paralleled GPU implementation of ROI Pooling and (2) by sharing part of the computation on the entire image. More specifically, we use a shared backbone network to extract high level features of the image and build the base detector and DCR V2 on top of the shared feature extractor. This design is based on the assumption that early stages of deep convolutional neural networks mainly extract low-level features (\eg edges and textures) and we assume these low-level features can be shared among different tasks. The base detector is the same as that in the Faster RCNN and DCR V2 module is a \textbf{deep} convolutional classifier on regional features which are pooled by ROI Pooling.}

\newcontent{To avoid the problem of feature sharing of classification and localization, we use a very deep residual network as the feature extractor of DCR V2. By placing this network on top of regional feature, the DCR V2 module is capable of learning translation invariance features for classifying regions. To introduce adaptive receptive field, we place the DCR V2 module at the early stage (by early stage, we mean layers that is close to input image). The early stages of network can learn texture-aware features that can be shared among different tasks as it has small stride and local receptive field. In our experiments, we also find it is beneficial to place DCR V2 module at early stage of the network.}

% \subsubsection{\newcontent{Efficient Inference of DCR V2}}
\newcontent{Although sharing part of the regional feature extraction reduces the computation, we still need to process a large batch of ROIs during inference which requires a lot of memory and computation. To further reduce the inference time, we propose a simple yet efficient strategy called ``top-sampling''. We find that the higher confidence score a false positive has, the larger impact it has (Fig. \ref{fig:motivation}), thus we place different importance on false positives based on their confidence scores. In ``top-sampling'', we only sample part of the detections whose confidence scores are within the top $p$ percent which can further reduce the inference time while preserving the accuracy. For example, if we choose $p=50\%$, then during the inference, we first score detections based on their maximum softmax scores and only pass top $50\%$ boxes into the DCR V2 module.}

\newcontent{The ``top-sampling'' strategy is also based on the fact that hard false positives that cannot be suppressed by post-processing (\eg NMS) usually have high confidence, while false positives with low confidence are less likely to be suppressed. In this way, we can achieve a speed-accuracy trade-off by ``attending'' to detections with high confidence scores. Experimentally, we demonstrate that processing only the top $50\%$ boxes degrades performance by less than $0.5\%$ while nearly halving the actual run time.}

\subsection{An AP interpretation of DCR}
The success of DCR V1~\cite{cheng2018revisiting} shows the necessity of removing false positives in object detection. However, one question in our mind is that \textbf{is the effectiveness of DCR comes from model ensemble?} The answer is \textbf{no} as we will show in Section~\ref{exps}. If that is the case, then what makes DCR so effective? We try to answer this question from the perspective of average precision (AP). To calculate AP, detections are first sorted according to their confidence scores which makes false positives in earlier ranking impact more than false positives in later ranking. Although DCR cannot remove all false positives, it re-ranks detections to make false positives appear later than true positive. The recent success of Mask Scoring R-CNN~\cite{huang2019mask} can be served as a strong evidence that ranking matters.

% \vspace{-3mm}
\subsection{Training}
\label{training}

\subsubsection{Training DCR V1}
Since there is no error propagates from the DCR module to Faster RCNN, we train our object detector in a two-step manner. First, we train Faster RCNN to converge. Then, we train our DCR module on mini-batches sampled from hard false positives of Faster RCNN. Parameters of DCR module are pre-trained by ImageNet dataset \cite{deng2009imagenet}. We follow the image-centric method \cite{girshick2015fast} to sample $N$ images with a total mini-batch size of $R$ boxes, \ie $R/N$ boxes per image. We use $N=1$ and $R=32$ throughout experiments. We use a different sampling heuristic that we sample not only foreground and background boxes but also hard false positive \textbf{uniformly}. Because we do not want to apply any prior knowledge to impose unnecessary bias on classifier. However, we observed that boxes from the same image have little variance. Thus, we fix Batch Normalization layer with ImageNet training set statistics. The newly added linear classifier (fully connected layer) is set with 10 times of the base learning rate since we want to preserve translation invariance features learned on the ImageNet dataset.

% \vspace{-2mm}
\subsubsection{\newcontent{Training DCR V2}}

\newcontent{We train our DCR V2 module with the base detector in an end-to-end manner. The training of the base detector is the same as \cite{ren2015faster} and we mainly discuss the training of DCR V2 in detail. During training, we first sample $R$ boxes from the outputs of the bounding box regression branch in the base detector, then we use these boxes as ROIs to extract regional features in the DCR V2 module. And the training of DCR V2 module is simply minimizing the cross entropy loss. In our experiments, we follow the optimal sampling strategy and label assignment that is used for training DCR V1. The final loss term is:
$$\mathcal{L}=\mathcal{L}_{\text{RPN}}+\mathcal{L}_{\text{RCNN}}+\mathcal{L}_{\text{DCRV2}}$$}

\subsection{\newcontent{Difference with Other Cascade Methods}}
\newcontent{CRAFT \cite{yang2016craft} or Cascade RCNN \cite{cai2017cascade} can be categorized as using cascade classifiers to improve object detection, but they are essentially different from our approach. In particular:}
\begin{enumerate}
\item
\newcontent{\textbf{The motivations are different.} CRAFT is motivated by ``divide and conquer'' philosophy, it is designed to split one tasks into several small tasks. Cascade RCNN is motivated by the fact that previous training methods (use IOU 0.5 to assign foregrounds) are misaligned with testing metrics (AP with IOU 0.5-0.95). Our motivation is that Faster RCNN is making many classification errors and we want to improve it by correctly classifying those misclassified regions.}

\item \newcontent{\textbf{The architecture design principals are different.} Both CRAFT and Cascade RCNN have fixed receptive field and share the backbone features. DCR uses a novel ``adaptive receptive field'', we name it ``adaptive'' because the effective receptive field is always equal to the size of the correspondent object proposal. The classifier in DCR uses different feature from object detector.}

\item \newcontent{\textbf{The training and inference strategies are different.} All three method train the cascade classifier based on the output of the previous classifier. CRAFT uses \emph{all} output except background objects of the previous classifier and uses \emph{same} IOU threshold (\emph{i.e.} 0.5) to assign labels. Cascade RCNN uses \emph{all} output but reassign foreground labels based on \emph{different} IOU thresholds (\emph{i.e.} 0.5, 0.6, 0.7). DCR uses only \emph{part} of the previous output with a novel sampling heuristic (hard false positive sampling). In inference, both CRAFT and Cascade RCNN use prediction of the \emph{last} classifier, their hypothesis is the cascade classifier is always better than the previous one. However, DCR uses \emph{both} classifiers, our hypothesis is that our classifiers complement each other.}
\end{enumerate}

\begin{table*}[t]
	%\scriptsize
	\centering
	%\resizebox{1\textwidth}{!}{
	\begin{minipage}[t]{0.22\textwidth}
		\vspace{0pt}
		\centering
		\begin{tabular}{l|c}
			Sample method & mAP \\
			\hline
			Baseline & 79.8 \\
			\hline
			Random & 81.8\\ 
			FP Only & 81.4 \\
			FP+FG & 81.6 \\
			FP+BG & 80.3 \\
			FP+FG+BG & \textbf{82.3} \\
			RCNN-like & 81.7 \\	
		\end{tabular}
		%		\caption{(a)}
		%			\caption*{ (a) Ablation study on sampling heuristic.}
	\end{minipage}% <---- don't forget this %
	\begin{minipage}[t]{0.22\textwidth}
		\vspace{0pt}
		\centering	
		\begin{tabular}{l|c}
			FP Score & mAP \\
			\hline
			Baseline & 79.8 \\
			\hline
			0.20 & 82.2\\
			0.25 & 81.9 \\
			0.30 & \textbf{82.3} \\
			0.35 & 82.2 \\
			0.40 & 82.0 \\
		\end{tabular}
		%			\caption*{(b) Ablation study on threshold for hard false positive score}
	\end{minipage}	
	\begin{minipage}[t]{.22\textwidth}
		\vspace{0pt}
		\centering		
		\begin{tabular}{l|c}
			Sample size & mAP \\
			\hline
			Baseline & 79.8 \\
			\hline
			8 Boxes & 82.0 \\
			16 Boxes & 82.1 \\
			32 Boxes & \textbf{82.3} \\
			64 Boxes & 82.1 \\			
		\end{tabular}
		%			\caption*{(c) Ablation study on sampling number.}
	\end{minipage}% <---- don't forget this %
	\begin{minipage}[t]{.3\textwidth}
		\vspace{0pt}
		\begin{tabular}{l|c|c}
			ROI scale & mAP & Test Time \\
			\hline
			Baseline & 79.8 & 0.0855 \\
			\hline
			$56 \times 56$ & 80.6 & 0.0525\\
			$112 \times 112$ & 82.0 & 0.1454\\
			$224 \times 224$ & \textbf{82.3} & 0.5481\\
			$320 \times 320$ & 82.0 & 1.0465\\
		\end{tabular}
		%			\caption{Ablation study on box scale.}			
	\end{minipage}
	%	}	

	\begin{minipage}[b]{.22\textwidth}
		\centering
		(a)
	\end{minipage}
	\begin{minipage}[b]{.22\textwidth}
		\centering
		(b)
	\end{minipage}
	\begin{minipage}[b]{.22\textwidth}
		\centering
		(c)
	\end{minipage}
	\begin{minipage}[b]{.3\textwidth}
		\centering
		(d)
	\end{minipage}
	%	\resizebox{1\textwidth}{!}{		
	\begin{minipage}[t]{.35\textwidth}
		\vspace{0pt}
		\begin{tabular}{l|c|c} 
			DCR Depth & mAP & Test Time \\
			\hline
			Baseline & 79.8 & 0.0855  \\
			\hline
			18 & 81.4 & 0.1941 \\
			34 & 81.9 & 0.3144 \\
			50 & 82.3 & 0.5481 \\
			101 & 82.3 & 0.9570 \\
			152 & \textbf{82.5} & 1.3900 \\
		\end{tabular}
		%		\caption{Ablation study on classifier depth.}		
	\end{minipage}
	\begin{minipage}[t]{.22\textwidth}
		\vspace{0pt}
		\begin{tabular}{l|c}
			Base detector & mAP \\
			\hline
			Faster & 79.8 \\
			Faster+DCR & 82.3\\
			\hline
			DCN & 81.4 \\
			DCN+DCR & 83.2\\
		\end{tabular}
		%		\caption{Ablation study on base detector.}		
	\end{minipage}
	\begin{minipage}[t]{.35\textwidth}		
		\vspace{0pt}
		\begin{tabular}{l|c}
			Model capacity & mAP \\
			\hline
			Faster w/ Res101 & 79.8 \\
			Faster w/ Res152 & 80.3 \\
            Faster Ensemble & 81.1 \\
			Faster w/ Res101+DCR-50 & 82.3\\
		\end{tabular}
		%		\caption{Ablation study on model capacity.}
	\end{minipage}	
	
	\begin{minipage}[b]{.35\textwidth}
		\centering
		(e)
	\end{minipage}
	\begin{minipage}[b]{.22\textwidth}
		\centering
		(f)
	\end{minipage}
	\begin{minipage}[b]{.35\textwidth}
		\centering
		(g)
	\end{minipage}
	%	}
	
	\caption{Ablation studies results of DCR V1. Evaluate on PASCAL VOC2007 test set. Baseline is Faster RCNN with ResNet-101 as backbone. DCR V1 module uses ResNet-50. (a) Ablation study on sampling heuristics. (b) Ablation study on threshold for defining hard false positives. (c) Ablation study on sampling size. (d) Ablation study on ROI scale and test time (measured in seconds/image). (e) Ablation study on depth of DCR module and test time (measured in seconds/image). (f) DCR module with difference base detectors. Faster denotes Faster RCNN and DCN denotes Deformable Faster RCNN, both use ResNet-101 as backbone. (g) Comparison of Faster RCNN with same size as Faster RCNN + DCR.}
	\label{ablation}
    % \vspace{-4mm}
\end{table*}

% \vspace{-3mm}
\section{Experiments}
\label{exps}
% \vspace{-3mm}
\subsection{Implementation Details}

\subsubsection{DCR V1}
We train base detectors, \eg Faster RCNN, following their original implementations. We use default settings in \ref{training} for DCR module, we use ROI size $224 \times 224$ and use a threshold of 0.3 to identify hard false positives. Our DCR module is first pre-trained on the ILSVRC 2012 dataset \cite{deng2009imagenet}. In fine-tuning, we set the initial learning rate to $0.0001$ \emph{w.r.t}. one GPU and weight decay of $0.0001$. We follow linear scaling rule in \cite{goyal2017accurate} for data parallelism on multiple GPUs and use 4 GPUs for PASCAL VOC and 8 GPUs for COCO. Synchronized SGD with momentum $0.9$ is used as optimizer. No data augmentation except horizontal flip is used.

% \vspace{-2mm}
\subsubsection{\newcontent{DCR V2}}
\newcontent{The implementations of Faster RCNN is the same as their original ones and we will only discuss detail implementation of our DCR V2 module using a 101-layer ResNet as example.}

\begin{figure*}[ht]
	\centering
	\includegraphics[width=1.0\textwidth]{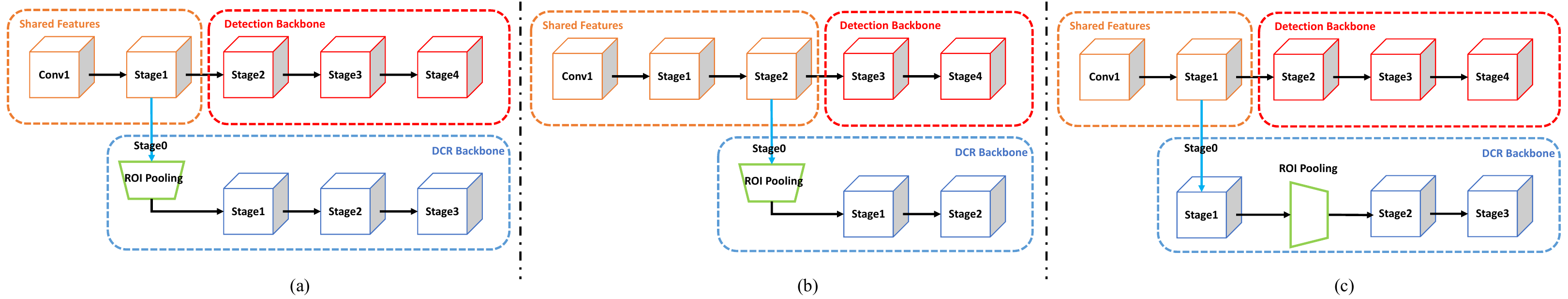}
	\caption{\newcontent{Illustration of ablation studies performed on DCR V2. For simplicity, detection head including RPN, ROI Pooling, cls and bbox regression are ignored. (a) Our default DCR V2 structure. (b) Ablation study on the amount of features to share. This is an example when DCR V2 is connected after Stage2 of ResNet with mAP 82.8\%. (c) Ablation study on adaptive receptive field of DCR V2. This is and example when ROI Pooling is placed after Stage1 of DCR V2 with mAP 82.9\%.}}
	\label{fig:network}
    %  \vspace{-4mm}
\end{figure*}

\newcontent{Fig \ref{fig:network} (a) shows a detailed block diagram of default DCR V2 module. Follow the naming convention in mainstream frameworks' implementation of ResNet, \eg MXNET and PyTorch, we denote the five stages of ResNet as conv1 (the first Convolution, BatchNorm and Max Pooling that reduce input resolution by a factor of 4), Stage1 (the first residual stage containing 3 residual blocks, with output stride 4), Stage2 (the second residual stage containing 4 residual blocks, with output stride 8), Stage3 (the third residual stage containing 23 residual blocks, with output stride 16) and Stage4 (the fourth residual stage containing 3 residual blocks, with output stride 32). Following \cite{lin2017feature}, we append an additional 3x3 convolution with 256 output channel after Stage4. We place the RPN at the end of Stage3 and the ROI Pooling that takes proposals from RPN as input at the end of Stage4. For the RCNN-head, we follow \cite{lin2017feature} to use two MLP with 1024 hidden layers followed by a classification branch and a bounding box regression branch. To construct the DCR V2 module, we place another ROI Pooling at the end of Stage1 that takes detections of RCNN-head as ROI input. On top of the ROI Pooling, we simply copy Stage2, Stage3, Stage4 of ResNet and add a global average pooling followed with a linear classifier. We initialize both Faster RCNN and DCR V2 module with same ImageNet pretrained weights.}

\newcontent{We use an initial learning rate of 0.0005 and a batchsize of 1 for each GPU. The weight decay is set to 0.0005 and momentum is set to 0.9. We only use horizontal flip during training.}

%%%%%%%%%%%%%%%%%%%%%%%%%%%%%%%%%%%%%
\begin{table*}[t]
	%\scriptsize
	\centering
	%\resizebox{1\textwidth}{!}{
	\begin{minipage}[t]{.3\textwidth}		
		\vspace{0pt}
		\begin{tabular}{l|c|c}
			DCR V2 Stage & mAP & Test Time\\
			\hline
			Baseline & 79.8 & 0.0855\\
            \hline
			Stage 4 & 78.4 &  0.0945\\
            Stage 3 & 80.0 &  0.1543\\
			Stage 2 & 82.8 &  0.6540\\
            Stage 1 & \textbf{83.0} & 0.7929\\
            \hline
            Stage 0 \tiny{(DCR V1)} & 82.3 & 0.9570
		\end{tabular}
		%		\caption{Ablation study on model capacity.}
	\end{minipage}	
	\begin{minipage}[t]{.35\textwidth}		
		\vspace{0pt}
		\begin{tabular}{l|c|c}
			ROIPooling Stage & mAP & Test Time \\
			\hline
			Baseline & 79.8 & 0.0855\\
            \hline
			Stage 3 (after DCR V2) & 79.8 & 0.1474\\
            Stage 2 & 80.7  & 0.2027\\
			Stage 1 & 82.9  & 0.6590\\
            Stage 0 (before DCR V2) & \textbf{83.0} & 0.7929\\
		\end{tabular}
		%		\caption{Ablation study on model capacity.}
	\end{minipage}	
	\begin{minipage}[t]{.3\textwidth}
		\vspace{0pt}
		\centering		
		\begin{tabular}{l|c|c}
			top-$p$ & mAP & Test Time \\
			\hline
% 			DCR V1 & 82.5 & 1.3900  \\
			DCR V1 & 82.3 & 0.9570  \\
			\hline
			100\% & 83.0 & 0.7929 \\
			75\% & 82.9 & 0.6323 \\
			50\% & 82.8 & 0.4653 \\
			25\% & 81.9 & 0.3015 \\
			0\% & 80.2 & 0.0855 \\
		\end{tabular}
		%			\caption*{(c) Ablation study on sampling number.}
	\end{minipage}% <---- don't forget this %

	\begin{minipage}[b]{.3\textwidth}
		\centering
		(a)
	\end{minipage}
	\begin{minipage}[b]{.35\textwidth}
		\centering
		(b)
	\end{minipage}
	\begin{minipage}[b]{.3\textwidth}
		\centering
		(c)
	\end{minipage}

	\caption{\newcontent{Ablation studies results of DCR V2 Module on Pascal VOC. Evaluate on PASCAL VOC2007 test set. Baseline is Faster RCNN with ResNet-101 as backbone. (a) Ablation study on the amount of feature sharing by adding DCR V2 branch after different stages in ResNet-101 backbone, ROI Pooling is placed before DCR V2 module. (b) Ablation study on adaptive receptive field by placing ROI Pooling after different stages of DCR V2 module, DCR V2 module is placed after Stage 1 of ResNet-101. (c) DCR V2 Top-Sampling during inference. Baseline model is Faster RCNN with ResNet-101 backbone. For the DCR V2 module, we place it at the end of Stage 1. The speed is measured on a single NVIDIA GTX 1080 TI.}}
	\label{dcrv2_ablation}
%     \vspace{-8mm}
\end{table*}

%%%%%%%%%%%%%%%%%%%%%%%%%%%%%%%%%%%%%
\begin{table}[]
    \centering
    \resizebox{0.49\textwidth}{!}{
    \begin{tabular}{lcccccc}
      Ablation Settings & AP & $\text{AP}_{50}$ & $\text{AP}_{75}$ & $\text{AP}_{\text{S}}$ & $\text{AP}_{\text{M}}$ & $\text{AP}_{\text{L}}$ \\
      \midrule \midrule
      \multicolumn{7}{l}{(a) Sampling Methods:} \\
      Baseline & 25.5 & 44.7 & 26.0 & 6.9 & 26.8 & 42.7 \\ 
      \midrule
	  Random & 31.3 & 54.1 & 32.3 & 12.1 & \textbf{34.3} & 49.2\\
      RCNN-like & \textbf{31.5} & 53.8 & \textbf{32.4} & \textbf{12.7} & 34.2 & \textbf{49.5}\\
	  DCR V1-like & 31.4 & \textbf{54.2} & 32.0 & 12.6 & 34.2 & 48.8\\
	  \midrule \midrule
	  \multicolumn{7}{l}{(b) Ensemble Methods:} \\
	  RCNN & 15.4 & 42.3 & 7.3 & 8.5 & 20.4 & 21.8 \\
	  Faster & 25.5 & 44.7 & 26.0 & 6.9 & 26.8 & 42.7\\
      \midrule
	  \{Faster, RCNN\} & 16.4 & 43.7 & 8.5 & 8.9 & 21.8 & 23.1\\
      Faster + RCNN & 28.4 & 51.3 & 28.3 & 11.1 & 30.6 & 45.8 \\
	  DCR V2 & \textbf{31.4} & \textbf{54.2} & \textbf{32.0} & \textbf{12.6} & \textbf{34.2} & \textbf{48.8}\\
	  \midrule \midrule
	  \multicolumn{7}{l}{(c) Bbox Regression:} \\
      Baseline & 25.5 & 44.7 & 26.0 & 6.9 & 26.8 & 42.7\\
	  \midrule
	  DCR V2 w/ bbox reg & 31.2 & 53.9 & 31.9 & 12.3 & \textbf{34.3} & 48.4\\
	  DCR V2 w/o bbox reg & \textbf{31.4} & \textbf{54.2} & \textbf{32.0} & \textbf{12.6} & 34.2 & \textbf{48.8}\\
	  \midrule \midrule
	  \multicolumn{7}{l}{(d) Score Fusion Methods:} \\
      Baseline & 25.5 & 44.7 & 26.0 & 6.9 & 26.8 & 42.7\\
	  \midrule
	  DCR V2 score only & 28.9 & 50.4 & 29.2 & 12.1 & 31.7 & 44.9\\
	  DCR V2 + Faster, multiply & 31.4 & \textbf{54.2} & 32.0 & 12.6 & 34.2 & \textbf{48.8}\\
	  DCR V2 + Faster, average & \textbf{31.6} & \textbf{54.2} & \textbf{32.3} & \textbf{13.3} & \textbf{34.4} & 48.7\\
    \end{tabular}}
    \caption{\revision{Ablation studies results of DCR V2 Module on COCO. Evaluate on COCO2017 val set. Baseline is Faster RCNN with ResNet-50 as backbone. (a) Ablation study on sampling method. (b) Ablation study on ensemble RCNN with Faster RCNN. (c) Ablation study on adding bbox regression to DCR. (d) Ablation study on score fusion method}}
    \label{dcrv2_coco_ablation}
\end{table}

\subsection{\revision{Ablation Studies of DCR V1}}
% \vspace{-2mm}

We comprehensively evaluate our method on the PASCAL VOC detection benchmark \cite{Everingham10}. We use the union of VOC 2007 trainval and VOC 2012 trainval as well as their horizontal flip as training data and evaluate results on the VOC 2007 test set. We primarily evaluate the detection mAP with IoU 0.5 (mAP@0.5). Unless otherwise stated, all ablation studies are performed with ResNet-101 for Faster RCNN, 
ResNet-50 as classifier for our DCR V1 module and ResNet-101 for DCR V2 module.
% \vspace{-3mm}
\subsubsection{Ablation study on sampling heuristic}
We compare results with different sampling heuristic in training DCR module:
% \vspace{-2mm}
\begin{itemize}
\item 
random sample: a minibatch of ROIs are randomly sampled for each image
\item
hard false positive only: a minibatch of ROIs that are hard positives are sampled for each image
\item
hard false positive and background: a minibatch of ROIs that are either hard positives or background are sampled for each image
\item
hard false positive and foreground: a minibatch of ROIs that are either hard positives or foreground are sampled for each image
\item
hard false positive, background and foreground: the difference with random sample heuristic is that we ignore easy false positives during training.
\item
RCNN-like: we follow the Fast RCNN's sampling heuristic, we sample two images per GPU and 64 ROIs per image with \emph{fg}:\emph{bg}=1:3.
\end{itemize}
% \vspace{-2mm}

Results are shown in Table \ref{ablation} (a). We find that the result is insensitive to sampling heuristic. Even with random sampling, an improvement of 2.0\% in mAP is achieved. With only hard false positive, the DCR achieves an improvement of 1.6\% already. Adding foreground examples only further gains a 0.2\% increase. Adding background examples to false negatives harms the performance by a large margin of 1.1\%. We hypothesize that this is because comparing to false positives, background examples dominating in most images results in a classifier bias to predicting background. This finding demonstrate the importance of hard negative in DCR training. Unlike RCNN-like detectors, we do not make any assumption of the distribution of hard false positives, foregrounds and backgrounds. To balance the training of classifier, we simply uniformly sample from the union set of hard false positives, foregrounds and backgrounds. This uniform sample heuristic gives the largest gain of 2.5\% mAP. We also compare our training with RCNN-like training. Training with RCNN-like sampling heuristic with \emph{fg}:\emph{bg}=1:3 only gains a margin of 1.9\%.
% \vspace{-3mm}
\subsubsection{Ablation study on other hyperparameters}

We compare results with different threshold for defining hard false positive: [0.2, 0.25, 0.3, 0.35, 0.4]. Results are shown in Table \ref{ablation} (b). We find that the results are quite insensitive to threshold of hard false positives and we argue that this is due to our robust uniform sampling heuristic. With hard false positive threshold of 0.3, the performance is the best with a gain of 2.5\%.

We also compare the influence of size of sampled RoIs during training: [8, 16, 32, 64]. Results are shown in Table \ref{ablation} (c). Surprisingly, the difference of best and worst performance is only 0.3\%, meaning our method is highly insensitive to the sampling size. With smaller sample size, the training is more efficient without severe drop in performance.

% \vspace{-3mm}
\subsubsection{Speed and accuracy trade-off}
There are in general two ways to reduce inference speed, one is to reduce the size of input and the other one is to reduce the depth of the network. We compare 4 input sizes: $56\times56$, $112\times112$, $224\times224$, $320\times320$ as well as 5 depth choices: 18, 34, 50, 101, 152 and their speed. Results are shown in Table \ref{ablation} (d) and (e). The test speed is linearly related to the area of input image size and there is a severe drop in accuracy if the image size is too small, \eg $56\times56$. For the depth of classifier, deeper model results in more accurate predictions but also more test time. We also notice that the accuracy is correlated with the classification accuracy of classification model, which can be used as a guideline for selecting DCR module.

% \vspace{-3mm}
\subsubsection{Generalization to more advanced object detectors}
We evaluate the DCR V1 module on Faster RCNN and advanced Deformable Convolution Nets (DCN) \cite{dai2017deformable}. Results are shown in Table \ref{ablation} (f). Although DCN is already among one of the most accurate detectors, its classifier still produces hard false positives and our proposed DCR module is effective in eliminating those hard false positives. More results of DCR V2 on advanced detectors are shown in Table \ref{tab:voc07}, \ref{tab:voc12}, \ref{tab:coco-minival} and \ref{tab:coco-test}.
% \vspace{-3mm}
\subsubsection{Where is the gain coming from?}
One interesting question is where the accuracy gain comes from. Since we add a large convolutional network on top of the object detector, does the gain simply comes from more parameters? Or, is DCR an ensemble of two detectors?
To answer this question, we compare the results of Faster RCNN with ResNet-152 as backbone (denoted Faster-152) and Faster RCNN with ResNet-101 backbone + DCR-50 (denoted Faster-101+DCR-50) and results are shown in Table \ref{ablation} (g). Since the DCR module is simply a classifier, the two network have approximately the same number of parameters. However, we only observe a marginal gain of 0.5\% with Faster-152 while our Faster-101+DCR-50 has a much larger gain of 2.5\%. To show DCR is not simply then ensemble to two Faster RCNNs, we further ensemble Faster RCNN with ResNet-101 and ResNet-152 and the result is 81.1\% which is still 1.1\% worse than our Faster-101+DCR-50 model. This means that the capacity does not merely come from more parameters or ensemble of two detectors.

\subsection{\revision{Ablation Studies of DCR V2}}

\subsubsection{\revision{Ablation study on general design}}
\revision{We perform two ablation studies to validate general design choices for our DCR V2 module on Pascal VOC dataset: 1. after which stage to add DCR V2 module in the backbone? 2. where to add ROI Pooling to extract regional feature in the DCR V2 module? Results are shown in Table~\ref{dcrv2_ablation} (a) and (b).}

\revision{We experiment on adding DCR V2 \textbf{after} Stage1, Stage2, Stage3, Stage4 in ResNet-101 to study the effect of feature sharing between Faster RCNN and DCR V2, and we simply use the remaining stage(s) that are not shared for DCR V2 module (Fig \ref{fig:network} (b)). In this case, ROI Pooling for DCR V2 is placed before DCR V2 module, \ie DCR V2 module acts directly on the regional feature. As we place DCR V2 to earlier stages, we let Faster RCNN share less features with DCR V2 module and we observe an increase in the performance (mAP is increased from 78.4 to 83.0 as we place DCR V2 from Stage4 to Stage1). However, the inference time increases as expected. These results are consistent with our hypothesis in Section \ref{problem1} that feature sharing between classification and localization might be harmful.}

\revision{When the position of DCR V2 module is fixed (after Stage1 of the ResNet), we also explore the effect of ROI Pooling position within DCR V2 (Fig \ref{fig:network} (c)). The position of ROI Pooling decides a trade-off between computing regional feature and image-level feature. If we place ROI Pooling at Stage3 (the last stage) of DCR V2 (after the module), all features are computed at image level and the mAP is only 79.8. If we place ROI Pooling at Stage0 (before DCR V2 module), then all features of DCR V2 are computed at ROI level, the mAP is 83.0. This results are consistent with our hypothesis in Section \ref{problem3} that the model should have adaptive receptive field, which means, placing ROI Pooling at earlier stage is beneficial.}

\subsubsection{\revision{Speed and accuracy trade-off}}
\revision{Results of our DCR V2 with different running times are shown in Table~\ref{dcrv2_ablation} (c). We evaluate the running time with DCR V2 after Stage1 on a single NVIDIA GTX 1080 TI GPU. Comparing with DCR V1 which has a run time of 1.3900 seconds/image, DCR V2 only requires 0.7929 seconds/image to get even better performance (+0.5\% mAP) which is 1.75 times faster than DCR V1. If we use the ``top-sampling'' strategy to only sample top $50\%$ of the detections, the run time becomes 0.4653 seconds/image with a degradation of 0.2\% in mAP, which is 3 times faster than DCR V1 and 1.7 times faster than DCR V2 with out sampling. Moreover, as mentioned earlier, the actual run time is much longer than 1.3900 seconds/image as we do not count the running time to crop and resize 300 images on the original image. That is, DCR V2 has a much better speed/accuracy trade-off over DCR V1.}

\begin{table*}[tb]\setlength{\tabcolsep}{1pt}
	\centering
    %\footnotesize
	%\tiny
    \resizebox{0.95\textwidth}{!}{
\begin{tabular}{l|c|cccccccccccccccccccc}
 Method  & mAP & \rotatebox{90}{aero} & \rotatebox{90}{bike} & \rotatebox{90}{bird} & \rotatebox{90}{boat} & \rotatebox{90}{bottle} & \rotatebox{90}{bus} & \rotatebox{90}{car} & \rotatebox{90}{cat} & \rotatebox{90}{chair} & \rotatebox{90}{cow} & \rotatebox{90}{table} & \rotatebox{90}{dog} & \rotatebox{90}{horse} & \rotatebox{90}{mbike} & \rotatebox{90}{person} & \rotatebox{90}{plant} & \rotatebox{90}{sheep} & \rotatebox{90}{sofa} & \rotatebox{90}{train} & \rotatebox{90}{tv} \\
        \hline

      Faster~\cite{he2016deep} 
      & 76.4 & 79.8 & 80.7 & 76.2 & 68.3 & 55.9 & 85.1 & 85.3 & 89.8 & 56.7 & 87.8 & 69.4 & 88.3 & 88.9 & 80.9 & 78.4 &  41.7 & 78.6 & 79.8 & 85.3 & 72.0\\
     
     R-FCN~\cite{dai2016r} & 80.5 & 79.9 & 87.2 & 81.5 & 72.0 & 69.8 & 86.8 & 88.5 & 89.8 & 67.0 & 88.1 &  74.5 & 89.8 & \textbf{90.6} & 79.9 & 81.2 & 53.7 & 81.8 & 81.5 & 85.9 & 79.9\\
     
     SSD~\cite{liu2016ssd,fu2017dssd} & 80.6 & 84.3 & 87.6 & \textbf{82.6} & 71.6 & 59.0 & 88.2 & 88.1 & 89.3 & 64.4 & 85.6 & 76.2 & 88.5 & 88.9 & 87.5 & 83.0 & 53.6 & 83.9 & 82.2 & 87.2 & 81.3 \\
     
      DSSD~\cite{fu2017dssd} 
      & 81.5 & 86.6 & 86.2 & \textbf{82.6} & 74.9 & 62.5 &89.0 & 88.7 & 88.8 & 65.2 & 87.0 & 78.7 & 88.2 & 89.0 & 87.5 & 83.7 & 51.1 & 86.3 & 81.6 & 85.7 & 83.7 \\

\hline
     Faster (2fc)& 79.8 & 79.6 & 87.5 & 79.5 & 72.8 & 66.7 & 88.5 & 88.0 & 88.9 & 64.5 & 84.8 & 71.9 & 88.7 & 88.2 & 84.8 & 79.8 & 53.8 & 80.3 & 81.4 & 87.9 & 78.5\\
     Faster-DCR V1 & 82.5 & 80.5 & \textbf{89.2} & 80.2 & 75.1 & 74.8 & 79.8 & 89.4 & 89.7 & 70.1 & 88.9 & 76.0 & 89.5 & 89.9 & 86.9 & 80.4 & 57.4 & 86.2 & 83.5 & 87.2 & 85.3\\
     \newcontent{Faster-DCR V2} & 83.0 & 87.7 & 88.3 & 80.9 & 77.1 & 73.6 & 90.0 & 89.1 & 90.2 & 69.5 & 89.0 & 76.1 & 89.6 & 90.0 & \textbf{88.6} & 80.6 & 56.2 & 86.1 & 84.3 & \textbf{88.2} & 85.7\\
     
     \hline
     DCN (2fc) & 81.4 & 83.9 & 85.4 & 80.1 & 75.9 & 68.8 & 88.4 & 88.6 & 89.2 & 68.0 & 87.2 & 75.5 & 89.5 & 89.0 & 86.3 & 84.8 & 54.1 & 85.2 & 82.6 & 86.2 & 80.3\\
     DCN-DCR V1 & 84.0 & 89.3 & 88.7 & 80.5 & \textbf{77.7} & \textbf{76.3} & \textbf{90.1} & \textbf{89.6} & 89.8 & \textbf{72.9} & \textbf{89.2} & 77.8 & 90.1 & 90.0 & 87.5 & 87.2 & 58.6 & \textbf{88.2} & 84.3 & 87.5 & 85.0\\
%     \textcolor{red}{DCN-DCR V2} & 83.6 & 88.6 & 89.5 & 80.3 & 77.1 & 75.4 & 89.9 & 89.0 & 89.9 & 71.5 & 89.6 & 76.0 & 89.6 & 90.0 & 89.0 & 87.1 & 56.8 & 85.9 & 84.6 & 88.0 & 85.0\\
    \newcontent{DCN-DCR V2} & \textbf{84.2} & \textbf{90.2} & 89.0 & 80.7 & \textbf{77.7} & 75.4 & 89.5 & \textbf{89.6} & \textbf{90.5} & 72.2 & 89.1 & \textbf{80.8} & \textbf{90.3} & 90.1 & 87.1 & \textbf{87.6} & \textbf{58.9} & 86.4 & \textbf{84.9} & \textbf{88.2} & \textbf{87.0}\\

		\end{tabular}
        }
		\caption{\textbf{PASCAL VOC2007 \texttt{test} detection results.}}% \textcolor{red}{DCN-DCR V2 with mAP 84.2 is achieved by adding  DCR V2 after stage1 of ResNet-101 and add ROIPooling at stage1 of DCR module.}}
\label{tab:voc07}
% \vspace{-8mm}
\end{table*}

\begin{table*}[htb]\setlength{\tabcolsep}{1pt}
	\centering
	%\tiny
    \resizebox{0.95\textwidth}{!}{
\begin{tabular}{l|c|cccccccccccccccccccc}
 Method  & mAP & \rotatebox{90}{aero} & \rotatebox{90}{bike} & \rotatebox{90}{bird} & \rotatebox{90}{boat} & \rotatebox{90}{bottle} & \rotatebox{90}{bus} & \rotatebox{90}{car} & \rotatebox{90}{cat} & \rotatebox{90}{chair} & \rotatebox{90}{cow} & \rotatebox{90}{table} & \rotatebox{90}{dog} & \rotatebox{90}{horse} & \rotatebox{90}{mbike} & \rotatebox{90}{person} & \rotatebox{90}{plant} & \rotatebox{90}{sheep} & \rotatebox{90}{sofa} & \rotatebox{90}{train} & \rotatebox{90}{tv} \\

        \hline
        Faster~\cite{he2016deep} & 73.8 & 86.5 & 81.6 & 77.2 & 58.0 & 51.0 & 78.6 & 76.6 & 93.2 & 48.6 & 80.4 & 59.0 & 92.1 & 85.3 & 84.8 & 80.7 & 48.1 & 77.3 & 66.5 & 84.7 & 65.6 \\

       R-FCN~\cite{dai2016r} &  77.6 & 86.9 & 83.4 & 81.5& 63.8& 62.4 & 81.6 & 81.1 & 93.1 & 58.0 & 83.8 & 60.8& 92.7 & 86.0 & 84.6 & 84.4 & 59.0 & 80.8 & 68.6& 86.1 & 72.9 \\

	SSD~\cite{liu2016ssd,fu2017dssd} & 79.4 & 90.7 & 87.3 & 78.3 & 66.3 & 56.5 & 84.1 & 83.7 &  94.2 & 62.9 & 84.5 & 66.3 & 92.9 & 88.6 & 87.9 & 85.7 & 55.1 & 83.6  & \textbf{74.3} & 88.2 & 76.8 \\
   
      DSSD~\cite{fu2017dssd} & 80.0 & \textbf{92.1} & 86.6 & 80.3 & 68.7 & 58.2 & 84.3 & \textbf{85.0} & \textbf{94.6} & 63.3 & 85.9 & 65.6 & 93.0 & 88.5 & 87.8 & 86.4 & 57.4 & 85.2 & 73.4 & 87.8 & 76.8  \\
     
     \hline
     Faster (2fc)& 77.3 & 87.3 & 82.6 & 78.8 & 66.8 & 59.8 & 82.5 & 80.3 & 92.6 & 58.8 & 82.3 & 61.4 & 91.3 & 86.3 & 84.3 & 84.6 & 57.3 & 80.9 & 68.3 & 87.5 & 71.4\\
     Faster-DCR V1 & 79.9 & 89.1 & 84.6 & 81.6 & 70.9 & 66.1 & \textbf{84.4} & 83.8 & 93.7 & 61.5 & 85.2 & 63.0 & 92.8 & 87.1 & 86.4 & 86.3 & 62.9 & 84.1 & 69.6 & 87.8 & \textbf{76.9}\\
     \newcontent{Faster-DCR V2} & 79.9 & 88.0 & 86.2 & 81.2 & 70.5 & 64.0 & 83.8 & 83.9 & 94.2 & 63.1 & 86.0 & 62.9 & 92.9 & 88.1 & 88.4 & 86.4 & 61.5 & 84.5 & 70.9 & 86.4 & 74.7\\

     \hline
     DCN (2fc) & 79.4 & 87.9 & 86.2 & 81.6 & 71.1 & 62.1 & 83.1 & 83.0 & 94.2 & 61.0 & 84.5 & 63.9 & 93.1 & 87.9 & 87.2 & 86.1 & 60.4 & 84.0 & 70.5 & \textbf{89.0} & 72.1\\
     DCN-DCR V1 & \textbf{81.2} & 89.6 & 86.7 & 83.8 & 72.8 & \textbf{68.4} & 83.7 & \textbf{85.0} & 94.5 & \textbf{64.1} & \textbf{86.6} & 66.1 & \textbf{94.3} & 88.5 & \textbf{88.5} & 87.2 & \textbf{63.7} & \textbf{85.6} & 71.4 & 88.1 & 76.1\\
     \newcontent{DCN-DCR V2} & 81.1 & 89.3 & \textbf{88.5} & \textbf{83.9} & \textbf{73.8} & 66.3 & 84.0 & \textbf{85.0} & 94.2 & 64.0 & 85.5 & \textbf{67.1} & 92.8 & \textbf{89.0} & 88.0 & \textbf{87.4} & 63.3 & 85.2 & 71.5 & 88.7 & 75.1\\

		\end{tabular}
        }
		\caption{\textbf{PASCAL VOC2012 \texttt{test} detection results.} }
\label{tab:voc12}
%  \vspace{-4mm}
\end{table*}

\begin{table}[tb]\setlength{\tabcolsep}{1pt}
	\centering
% 	\tiny
    \resizebox{0.49\textwidth}{!}{
\begin{tabular}{l|l|ccc|ccc}
 Method & Backbone & AP & $\text{AP}_{50}$ & $\text{AP}_{75}$ & $\text{AP}_{\text{S}}$ & $\text{AP}_{\text{M}}$ & $\text{AP}_{\text{L}}$ \\
        \hline
        
     Faster (2fc) & ResNet-101& 30.0 & 50.9 & 30.9 & 9.9 & 33.0 & 49.1 \\
     Faster-DCR V1 & ResNet-101 + ResNet-152 & 33.1 & 56.3 & 34.2 & 13.8 & 36.2 & 51.5 \\
     \newcontent{Faster-DCR V2} & ResNet-101 & 33.6 & 56.7 & 34.7 & 13.5 & 37.1 & 52.2 \\
     
     \hline
     DCN (2fc) & ResNet-101 & 34.4 & 53.8 & 37.2 & 14.4 & 37.7 & 53.1 \\
      DCN-DCR V1 & ResNet-101 + ResNet-152 & 37.2 & 58.6 & 39.9 & 17.3 & 41.2 & 55.5 \\
      \newcontent{DCN-DCR V2} & ResNet-101 & 37.5 & 58.6 & 40.1 & 17.2 & 42.0 & 55.5 \\
      
      \hline
     FPN & ResNet-101 & 38.2 & 61.1 & 41.9 & 21.8 & 42.3 & 50.3 \\
      FPN-DCR V1 & ResNet-101 + ResNet-152 & 40.2 & 63.8 & 44.0 & 24.3 & 43.9 & 52.6 \\
      \newcontent{FPN-DCR V2} & ResNet-101 & 40.3 & 62.9 & 43.7 & 24.3 & 44.6 & 52.7 \\
      
      \hline
     FPN-DCN & ResNet-101 & 41.4 & 63.5 & 45.3 & 24.4 & 45.0 & 55.1 \\
      FPN-DCN-DCR V1 & ResNet-101 + ResNet-152 & 42.6 & \textbf{65.3} & 46.5 & 26.4 & 46.1 & \textbf{56.4} \\
      \newcontent{FPN-DCN-DCR V2} & ResNet-101 & \textbf{42.8} & 65.1 & \textbf{46.8} & \textbf{27.1} & \textbf{46.6} & 56.1 \\

		\end{tabular}
        }
		\caption{\textbf{COCO2017 \texttt{val} detection results.}}
        % All detectors use ResNet-101 as backbone and DCR modules use ResNet-152 as base classifier.
\label{tab:coco-minival}
% \vspace{-2mm}
\end{table}
\begin{table*}[htb]\setlength{\tabcolsep}{1pt}
	\centering
% 	\tiny
    \resizebox{0.7\textwidth}{!}{
\begin{tabular}{l|l|ccc|ccc}
 Method & Backbone & AP & $\text{AP}_{50}$ & $\text{AP}_{75}$ & $\text{AP}_{\text{S}}$ & $\text{AP}_{\text{M}}$ & $\text{AP}_{\text{L}}$ \\
 \hline
%  YOLOv2~\cite{redmon2016yolo9000} & DarkNet-19 \cite{redmon2016yolo9000}
%   & 21.6 & 44.0 & 19.2 & 5.0 & 22.4 & 35.5 \\
%  SSD~\cite{liu2016ssd,fu2017dssd} & ResNet-101-SSD
%   & 31.2 & 50.4 & 33.3 & 10.2 & 34.5 & 49.8 \\
%  DSSD513~\cite{fu2017dssd} & ResNet-101-DSSD
%   & 33.2 & 53.3 & 35.2 & 13.0 & 35.4 & 51.1 \\
%  Faster+++~\cite{he2016deep} & ResNet-101-C4
%   & 34.9 & 55.7 & 37.4 & 15.6 & 38.7 & 50.9\\
%   G-RMI~\cite{huang2017speed} & Inception-ResNet-v2 \cite{szegedy2017inception}
%   & 34.7 & 55.5 & 36.7 & 13.5 & 38.1 & 52.0\\
%  FPN~\cite{lin2017feature} & ResNet-101-FPN
%   & 36.2 & 59.1 & 39.0 & 18.2 & 39.0 & 48.2\\
  Mask RCNN {\tiny{ICCV2017}}~\cite{he2017mask} & ResNeXt-101-FPN~\cite{xie2017aggregated}
  & 39.8 & 62.3 & 43.4 & 22.1 & 43.2 & 51.2 \\
 RetinaNet {\tiny{ICCV2017}}~\cite{lin2017focal} & ResNeXt-101-FPN
  & 40.8 & 61.1 & 44.1 & 24.1 & 44.2 & 51.2 \\
  Relation Net {\tiny{CVPR2018}}~\cite{hu2017relation} & ResNet-101
  & 39.0 & 58.6 & 42.9 & - & - & - \\
 Cascade RCNN {\tiny{CVPR2018}}~\cite{cai2017cascade} & ResNet-101-FPN
  & 42.8 & 62.1 & 46.3 & 23.7 & 45.5 & 55.2 \\
SNIP {\tiny{CVPR2018}}~\cite{singh2018analysis} & ResNet-101
  & 44.4 & 66.2 & 49.9 & 27.3 & 47.4 & 56.9 \\
  DetNet {\tiny{ECCV2018}}~\cite{Li_2018_ECCV} & DetNet-59-FPN
  & 40.3 & 62.1 & 43.8 & 23.6 & 42.6 & 50.0 \\
  CornerNet {\tiny{ECCV2018}}~\cite{Law_2018_ECCV} & Hourglass-104
  & 40.5 & 56.5 & 43.1 & 19.4 & 42.7 & 53.9 \\
  IOU-Net {\tiny{ECCV2018}}~\cite{Jiang_2018_ECCV} & ResNet-101-FPN
  & 40.6 & 59.0 & - & - & - & - \\
  ExtremeNet {\tiny{CVPR2019}}~\cite{zhou2019bottom} & Hourglass-104
  & 40.2 & 55.5 & 43.2 & 20.4 & 43.2 & 53.1 \\
  FSAF {\tiny{CVPR2019}}~\cite{zhu2019feature} & ResNeXt-64x4d-101-FPN
  & 42.9 & 63.8 & 46.3 & 26.6 & 46.2 & 52.7 \\
  FCOS {\tiny{ICCV2019}}~\cite{tian2019fcos} & ResNeXt-64x4d-101-FPN
  & 43.2 & 62.8 & 46.6 & 26.5 & 46.2 & 53.3 \\
  CenterNet511 {\tiny{ICCV2019}}~\cite{duan2019centernet} & Hourglass-104 
  & 44.9 & 62.4 & 48.1 & 25.6 & 47.4 & 57.4 \\
 
        \hline
        
     Faster (2fc) & ResNet-101 & 30.5 & 52.2 & 31.8 & 9.7 & 32.3 & 48.3 \\
     Faster-DCR V1 & ResNet-101 + ResNet-152 & 33.9 & 57.9 & 35.3 & 14.0 & 36.1 & 50.8 \\
     \newcontent{Faster-DCR V2} & ResNet-101 & 34.3 & 57.7 & 35.8 & 13.8 & 36.7 & 51.1 \\
     
     \hline
     DCN (2fc) & ResNet-101 & 35.2 & 55.1 & 38.2 & 14.6 & 37.4 & 52.6 \\
      DCN-DCR V1 & ResNet-101 + ResNet-152 & 38.1 & 59.7 & 41.1 & 17.9 & 41.2 & 54.7 \\
      \newcontent{DCN-DCR V2} & ResNet-101 & 38.2  & 59.7  & 41.2  & 17.3  & 41.7  & 54.6  \\
      
      \hline
     FPN & ResNet-101 & 38.8 & 61.7 & 42.6 & 21.9 & 42.1 & 49.7 \\
      FPN-DCR V1 & ResNet-101 + ResNet-152 & 40.7 & 64.4 & 44.6 & 24.3 & 43.7 & 51.9 \\
      \newcontent{FPN-DCR V2} & ResNet-101 & 40.8  & 63.6  & 44.5  & 24.3  & 44.3  & 52.0  \\
      
      \hline
     FPN-DCN & ResNet-101 & 41.7 & 64.0 & 45.9 & 23.7 & 44.7 & 53.4 \\
      FPN-DCN-DCR V1 & ResNet-101 + ResNet-152 & 43.1 & 66.1 & 47.3 & 25.8 & 45.9 & 55.3 \\
      \newcontent{FPN-DCN-DCR V2} & ResNet-101 & 43.5  & 65.9  & 47.6  & 25.8  & 46.6  & 55.9  \\

		\end{tabular}
        }
		\caption{\textbf{COCO \texttt{test-dev} detection results.} We only report results with single-scale test for a fair comparison.}
        % Comparison with state-of-the-arts reported in recent publications with different backbones.
\label{tab:coco-test}
%  \vspace{-4mm}
\end{table*}

\subsubsection{\revision{More ablation studies on COCO}}
To help us better understand why DCR V2 works well, we perform more ablation studies on COCO using ResNet-50 as the backbone by answering the following three questions:

\noindent \revision{ \textbf{(1) Does sampling method matter?} To answer the first question, we train DCR V2 with different sampling methods. Results are shown in Table~\ref{dcrv2_coco_ablation} (a). From the results, we can see that sampling methods influence DCR V2 less on COCO dataset, this might because COCO dataset is of larger scale than Pascal VOC dataset and using different sampling methods ends up similar amount of useful samples for DCR V2. But using sampling method from DCR V1 still gets the best $\text{AP}_{50}$ performance.}

\noindent \revision{\textbf{(2) Is DCR model simply a model ensemble?} DCR might look like an ensemble model at first glance, \textbf{but it is not a simple ensemble of Faster RCNN and RCNN and it performs better than the ensemble counter part.} To verify this, we train a RCNN model separately using RPN proposals. Results are shown in Table~\ref{dcrv2_coco_ablation} (b). The RCNN model only gets $\text{AP} = 15.4$, because we do not add bounding box regression branch for a fair comparison with DCR. However, RCNN still gets $\text{AP}_{50} = 42.3$ which is comparable with the results of Faster RCNN. We tried two ensemble methods. \{Faster, RCNN\} means applying NMS on detections from both Faster RCNN and RCNN and the ensemble model only gets $\text{AP} = 16.4$ and $\text{AP}_{50} = 43.7$. This means that directly combining detections from a strong detector and a weak detector does not have any gain. Next, we follows DCR V1 inference method to ensemble Faster RCNN and RCNN. That is, we pass results from Faster RCNN to RCNN and combine two confidence scores by multiplication. This ensemble methods offer a large improvement ending in $\text{AP} = 28.4$ and $\text{AP}_{50} = 51.3$, which improves Faster RCNN by an AP of 2.9. However, the results of DCR V2 is $\text{AP} = 31.4$ and $\text{AP}_{50} = 54.2$ which is much higher than the results of ensemble method which justifies that our DCR model is not a simple ensemble model.}

\noindent \revision{\textbf{(3) Is it necessary to decouple the localization task?} To show whether it is necessary to decouple both classification and localization tasks simultaneously, we also conduct additional experiment by adding one more bounding box regression branch to DCR V2. Results are shown in Table~\ref{dcrv2_coco_ablation} (c). We find that the new adding bounding box regression branch will not offer further improvement and slightly harms AP by 0.2. We consider the reason is that proposals from the first bounding box regression branch are already with good localization quality to recall most objects and thus the new localization branch is hard to bring new gains. Therefore, we no need to apply an additional decoupled bounding box refinement module for further refinement.}

\noindent \revision{\textbf{(4) Should scores be summed or muliplied?} For fusion method, we further perform ablation studies on three different score fusion method. By fusion method, we refer to how to assign final score to each detection. We use multiplication of DCR scores and Faster RCNN scores in DCR V1 \cite{cheng2018revisiting} and follow this method in DCR V2. Results are shown in Table~\ref{dcrv2_coco_ablation} (d). Using DCR V2 scores only already outperforms baseline Faster RCNN by a large margin. We find using average of two scores is slightly better by using multiplication. But we still use multiplication in rest of the experiments for consitency with DCR V1.}

% \vspace{-3mm}
\subsection{PASCAL VOC Results}
\subsubsection{VOC 2007}
We use a union of VOC2007 trainval and VOC2012 trainval as training set and test the model on VOC2007 test set. We use the default training setting and ResNet-152 as classifier for the DCR V1 and ResNet-101 for DCR V2 module. We train our model for 7 epochs and reduce learning rate by $\frac{1}{10}$ after 4.83 epochs. Results are shown in Table \ref{tab:voc07}. Notice that based on DCN as base detector, our single DCR module achieves competitive result of 84.2\% without using extra data (\eg COCO data), multi scale training/testing, ensemble or other post processing tricks.
% \vspace{-3mm}
\subsubsection{VOC 2012}
We use a union of VOC2007 trainvaltest and VOC2012 trainval as training set and test the model on VOC2012 test set. We use the same training setting of VOC2007. Results are shown in Table \ref{tab:voc12}. Based on DCN and DCR module, our model is the first model achieves over 81.0\% on the VOC2012 test set. A competitive result of 81.2\% is achieved using only single model, without any post processing tricks.
% \vspace{-3mm}

% \vspace{-2mm}
\subsection{COCO Results}
% \vspace{-1mm}
All experiments on COCO follow the default settings and use ResNet-152 for DCR V1 and ResNet-101 for DCR V2 module. We train our model for 8 epochs on the COCO dataset and reduce the learning rate by $\frac{1}{10}$ after 5.33 epochs. We report results on two different partition of COCO dataset. One partition is training on COCO2017 train with 115k images and evaluate results on the COCO2017 val with 5k images. The other partition is training on the standard COCO2017 trainval with 120k images and evaluate on the COCO test-dev by submitting results to the COCO evaluation server. We use Faster RCNN \cite{ren2015faster}, Feature Pyramid Networks (FPN) \cite{lin2017feature} and the Deformable ConvNets \cite{dai2017deformable} as base detectors.

% \vspace{-3mm}
\subsubsection{COCO2017 val}
Results are shown in Table \ref{tab:coco-minival}. DCR V2 consistently out-performs DCR V1. Our DCR V2 module improves Faster RCNN by 3.6\% from 30.0\% to 33.6\% in COCO AP metric. Faster RCNN with DCN is improved by 3.1\% from 34.4\% to 37.5\% and FPN is improved by 2.1\% from 38.2\% to 40.3\%. Notice that FPN+DCN is the base detector by top-3 teams in the COCO2017 detection challenge, but there is still an improvement of 1.4\% from 41.4\% to 42.8\%. This observation shows that currently there is no perfect detector that does not produce hard false positives.

\subsubsection{COCO test-dev}
Results are shown in Table \ref{tab:coco-test}. The trend is similar to that on the COCO2017 val, with Faster RCNN improved from 30.5\% to 34.3\%, Faster RCNN+DCN improved from 35.2\% to 38.2\%, FPN improved from 38.8\% to 40.8\% and FPN+DCN improved from 41.7\% to 43.5\%. We also compare our results with recent state-of-the-arts reported in publications and our best model achieves competitive result on COCO test-dev with ResNet as backbone. For a fair comparison, we only report results without any test time augmentation like multi-scale image pyramid and horizontal flip. Although there are recent methods (\eg CetnerNet511) outperforms DCR in AP, our DCR model still achieve state-of-the-art $\text{AP}_{50}$ performance. This is because DCR mainly focus on more precise classification instead of localization.

%  \vspace{-2mm}
\subsection{Discussions}

Our DCR module demonstrates extremely good performance in suppressing false positives. Fig \ref{fig:motivation} (a) compares total number of false positives on the VOC2007 test set. With our DCR module, the number of hard false is reduced by almost three times (orange). 
% The inference time of DCR module is proportional to the number of proposals, input size and network depth. Table \ref{ablation} (d), (e) compare the running time and the best model (DCR with depth 152) runs slower than the baseline Faster RCNN at the speed of 1.39 s/image on 1080 Ti GPU. However, this paper focuses more on the analysis of failure case of object detectors and accuracy boost, improvement to speed will be studied in the future.

\begin{figure*}[t]
	\centering
	\includegraphics[width=0.8\textwidth]{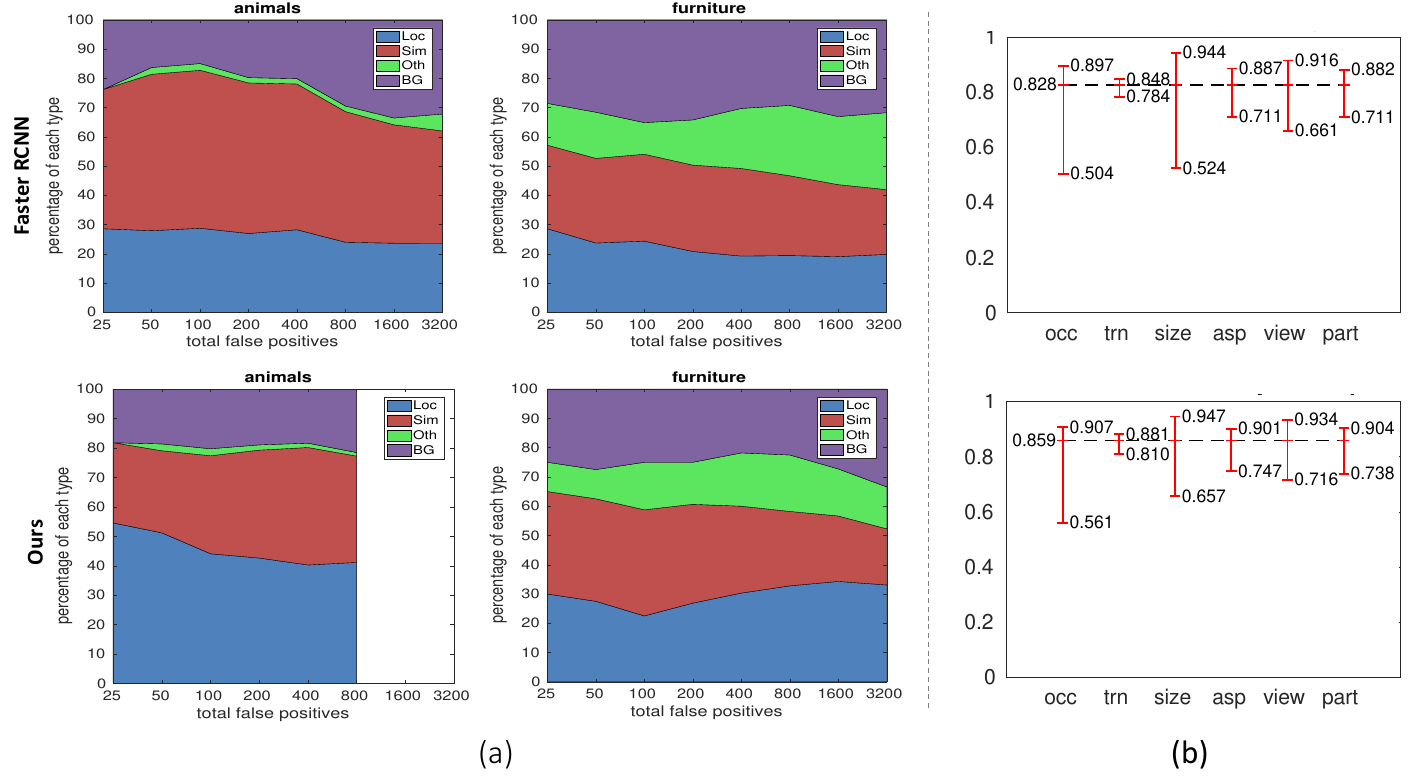}
	\caption{Analysis results between Faster RCNN (top row) and our methods (bottom row) by \cite{hoiem2012diagnosing}. Left of the dashed line: distribution of top false positive types. Right of the dashed line: sensitivity to object characteristics.}
	\label{fig:analysis}
\end{figure*}

\begin{figure*}
	\centering
	\includegraphics[width=1\textwidth]{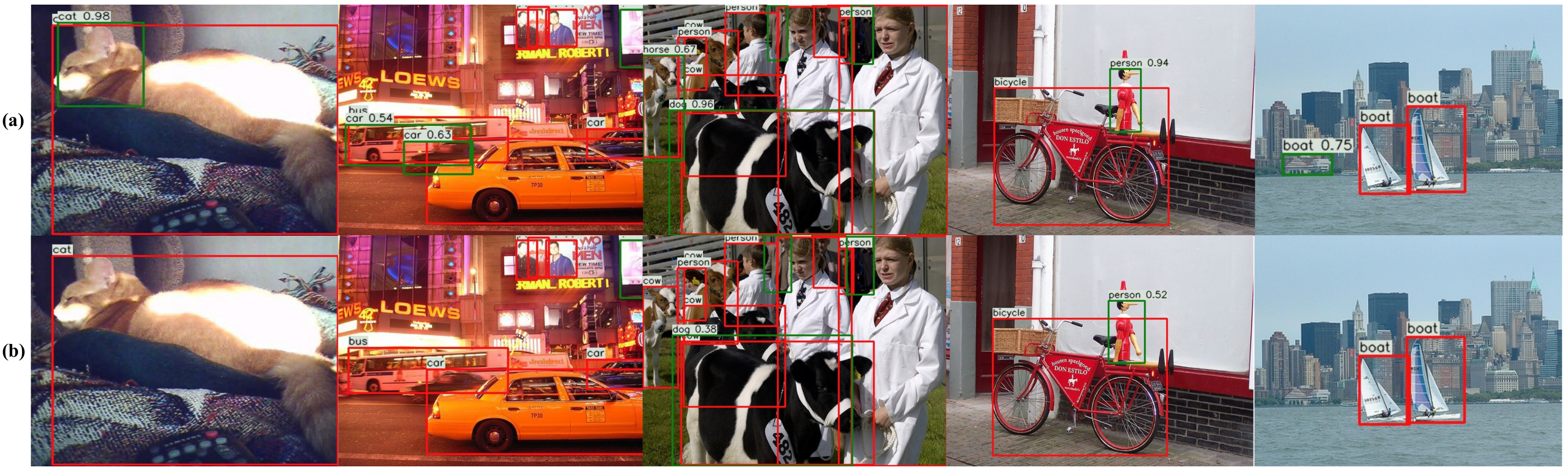}
	\caption{Comparison of hard false positives with confidence score higher than 0.3 between Faster RCNN and Our methods. Red box: ground truth object. Green box: hard false positive. Row (a), results from Faster RCNN. Row (b), results of our DCR module.}
	\label{fig:visualize_fp}
    % \vspace{-4mm}
\end{figure*}

\subsubsection{Error Analysis}
Following \cite{girshick2014rich}, we also use the detection analysis tool from \cite{hoiem2012diagnosing}, in order to gather more information of the error mode of Faster RCNN and DCR module. Analysis results are shown in Fig \ref{fig:analysis}. 

Fig \ref{fig:analysis} (a) shows the distribution of top false positive types as scores decrease. False positives are classified into four categories: (1) Loc: IOU with ground truth boxes is in the range of $[0.1, 0.5)$; (2) Sim: detections have at least 0.1 IOU with objects in predefined similar classes, \eg dog and cat are similar classes; (3) Oth: detections have at least 0.1 IOU with objects not in predefined similar classes; (4) BG: all other false positives are considered background. We observe that comparing with Faster RCNN, DCR module has much larger ratio of localization error and the number of false positives is greatly reduced on some classes, \eg in the animal class, the number of false positives is largely reduced by 4 times and initial percentage of localization error increases from less than 30\% to over 50\%. This statistics are consistent with motivations to reducing classification errors by reducing number of false positives. 

Fig \ref{fig:analysis} (b) compares the sensitivity of Faster RCNN and DCR to object characteristics. \cite{hoiem2012diagnosing} defines object with six characteristics: (1) occ: occlusion, where an object is occluded by another surface; (2) trn: truncation, where there is only part of an object; (3) size: the size of an object measure by the pixel area; (4) asp: aspect ratio of an object; (5) view: whether each side of an object is visible; (6) part: whether each part of an object is visible. Normalized AP is used to measure detectors performance and more details can be found in \cite{hoiem2012diagnosing}. In general, the higher the normalized AP, the better the performance. The difference between max and min value measure the sensibility of a detector, the smaller the difference, the less sensible of a detector. We observe that DCR improves normalized AP and sensitivity on all types of object and improves sensitivity significantly on occlusion and size. This increase came from the adaptive field of DCR, since DCR can focus only on the object area, making it less sensible to occlusion and size of objects.

\subsection{Visualization}
We visualize all false positives with confidence larger than 0.3 for both Faster RCNN and our DCR module in Fig \ref{fig:visualize_fp}. We observe that the DCR module successfully suppresses all three kinds of hard false positives to some extends. 

The first image shows reducing the first type of false positives (part of objects). Faster RCNN (top) classifies the head of the cat with a extremely high confidence (0.98) but it is eliminated by the DCR module. 

The second to the fourth images demonstrate situations of second type of false positives (similar objects) where most of false positives are suppressed (``car'' in the second image and ``horse'' in the third image). However, we find there still exists some limitations, \eg the ``dog'' in the third image where it is supposed to be a cow and the ``person'' in the fourth image. Although they are not suppressed, their scores are reduced significantly (0.96 $\rightarrow$ 0.38 and 0.92 $\rightarrow$ 0.52 respectively) which will also improve the overall performance. It still remains an open questions to solve these problems. We hypothesize that by using more training data of such hard false positives (\eg use data augmentation to generate such samples).

The last image shows example for the third type of false positives (backgrounds). A part of background near the ground truth is classified as a ``boat'' by the Faster RCNN and it is successfully eliminated by our DCR module.

% However, there are still problems with DCR module. We select some visualization of typical cases that cannot handled by DCR module in the third row in Fig \ref{fig:visualize_fp}. We find that part of object and extremely similar objects remain challenge situations to both Faster RCNN and DCR module, especially part of object with highly discriminative features. In the second image of the third row, we observe that DCR module can effectively reduce the confidence score of similar object by a large amount (0.92 $\rightarrow$ 0.52). In contrary, in the first image of the third row, it still cannot reduce score for strong discriminative parts (0.97 $\rightarrow$ 0.94). It still remains an open questions to solve these problems. We hypothesize that by using more training data of such hard false positives (\eg use data augmentation to generate such samples).

% \vspace{-5mm}
\section{Conclusion}
\label{conclusion}
% \vspace{-2mm}
In this paper, we analyze error modes of state-of-the-art region-based object detectors and study their potentials in accuracy improvement. We hypothesize that good object detectors should be designed following three principles: decoupled features, decoupled optimization and adaptive receptive field. Based on these principles, we propose a simple, effective and widely-applicable DCR module that achieves new state-of-the-art. \newcontent{Furthermore, we use experiments to support that a good detector should have decoupled features and adaptive receptive field and we observe an interesting novel trade-off between feature sharing and speed/accuracy.} In the future, we will further study what architecture makes a good object detector, adaptive feature representation in multi-task learning, and efficiency improvement of our DCR module. \newcontent{We hope this paper could motivate the community to study new directions of improving current object detection frameworks. Specifically, (1) DCR module still has 4.3\% mAP to be improved due to false positives and how to suppress these remaining false positives remains an open question. (2) how to design more efficient decoupled structure is yet another interesting research direction.}

%  \vspace{-4mm}
% \begin{acknowledgements}
% This work is in part supported by IBM-ILLINOIS Center for Cognitive Computing Systems Research (C3SR) - a research collaboration as part of the IBM AI Horizons Network; 
% %and by IARPA-BAA-16-13 Deep Intermodal Video Analytics (DIVA). 
% and by the Intelligence Advanced Research Projects Activity (IARPA) via Department of Interior/ Interior Business Center (DOI/IBC) contract number D17PC00341.
% The U.S. Government is authorized to reproduce and distribute reprints for Governmental purposes
% notwithstanding any copyright annotation thereon. Disclaimer: The views and
% conclusions contained herein are those of the authors and should not be interpreted
% as necessarily representing the official policies or endorsements, either
% expressed or implied, of IARPA, DOI/IBC, or the U.S. Government.
% We thank Jiashi Feng for helpful discussions.
% \end{acknowledgements}

%\begin{acknowledgements}
%\end{acknowledgements}


\bibliographystyle{spbasic}      % basic style, author-year citations
% \bibliography{}   % name your BibTeX data base
\bibliography{egbib}

% % Non-BibTeX users please use
% \begin{thebibliography}{}
% %
% % and use \bibitem to create references. Consult the Instructions
% % for authors for reference list style.
% % Format for Journal Reference
% Author, Article title, Journal, Volume, page numbers (year)
% Author, Book title, page numbers. Publisher, place (year)
% \end{thebibliography}

\end{document}